\documentclass[10pt,journal,compsoc]{IEEEtran}
\ifCLASSOPTIONcompsoc
  \usepackage[nocompress]{cite}
\else
  \usepackage{cite}
\fi
\ifCLASSINFOpdf
\else
\fi

\usepackage{graphicx}
\usepackage{amsmath,amssymb} % define this before the line numbering.
\usepackage{color}
\usepackage{sidecap}

\newcommand{\AND}{\mathrm{\scriptscriptstyle{And}}}
\newcommand{\OR}{\mathrm{\scriptscriptstyle{Or}}}

\newcommand{\argmax}{\operatornamewithlimits{argmax}}

\newcommand*{\TitleFont}{%
      \usefont{\encodingdefault}{\rmdefault}{b}{n}%
      \fontsize{20}{23}%
      \selectfont}

\begin{document}
{\title{\TitleFont Learning And-Or Model to Represent Context and Occlusion for Car Detection and Viewpoint Estimation}

\author{Tianfu~Wu$^*$, Bo~Li$^*$        
        and~Song-Chun~Zhu % <-this % stops a space
\IEEEcompsocitemizethanks{
	\IEEEcompsocthanksitem T.F. Wu is with the Department of Statistics, University of California, Los Angeles. 
	E-mail: tfwu@stat.ucla.edu
	\IEEEcompsocthanksitem B. Li is with Beijing Lab of Intelligent Information Technology, Beijing Institute of Technology, China and a visiting student at University of California, Los Angeles.  
E-mail: boli86@bit.edu.cn
\IEEEcompsocthanksitem S.-C. Zhu is with the Department of Statistics and Computer Science, University of California, Los Angeles.  
E-mail: sczhu@stat.ucla.edu
\IEEEcompsocthanksitem{$^*$ Joint first authors.}
}% <-this % stops an unwanted space

\thanks{Manuscript received MM DD, YYYY; revised MM DD, YYYY.}
}


\markboth{FOR REVIEW: IEEE TRANSACTIONS ON PATTERN ANALYSIS AND MACHINE INTELLIGENCE}%
{Shell \MakeLowercase{\textit{et al.}}: Learning And-Or Models to Represent Context and Occlusion for Car Detection and Viewpoint Estimation}

%\IEEEpubid{\makebox[\columnwidth]{\hfill 0000--0000/00/\$00.00~\copyright~2014 IEEE}%


\IEEEtitleabstractindextext{%
\begin{abstract}
This paper presents a method for learning an And-Or model to represent context and occlusion for car detection and viewpoint estimation. 
The learned And-Or model represents car-to-car context and occlusion configurations at three levels: (i) spatially-aligned cars, (ii) single car under different occlusion configurations, and (iii) a small number of parts.
The And-Or model embeds a grammar for representing large structural and appearance variations in a reconfigurable hierarchy. 
The learning process consists of two stages in a weakly supervised way (i.e., only bounding boxes of single cars are annotated). Firstly, the structure of the And-Or model is learned with three components: (a) mining multi-car contextual patterns based on layouts of annotated single car bounding boxes, (b) mining  occlusion configurations between single cars, and (c) learning different combinations of part visibility based on CAD simulations. The And-Or model is organized in a directed and acyclic graph which can be inferred by Dynamic Programming. Secondly, the model parameters (for appearance, deformation and bias) are jointly trained using Weak-Label Structural SVM. 
In experiments, we test our model on four car detection datasets --- the KITTI dataset \cite{Geiger12}, the PASCAL VOC2007 car dataset~\cite{pascal}, and two self-collected car datasets, namely the Street-Parking car dataset and the Parking-Lot car dataset, and three datasets for car viewpoint estimation --- the PASCAL VOC2006 car dataset~\cite{pascal}, the 3D car dataset~\cite{savarese}, and the PASCAL3D+ car dataset~\cite{xiang_wacv14}. Compared with state-of-the-art variants of deformable part-based models and other methods, our model achieves significant improvement consistently on the four detection datasets, and comparable performance on car viewpoint estimation.
\end{abstract}

}

\maketitle

\IEEEdisplaynontitleabstractindextext

% \fi
\IEEEpeerreviewmaketitle

%\vspace{-1mm}
\IEEEraisesectionheading{\section{Introduction}\label{sec:introduction}}

%\vspace{-2mm}
\subsection{Motivation and Objective}
%\IEEEPARstart{T}{he} recent literature of object detection has been focused on three aspects to improve accuracy performance: using hierarchical models such as discriminatively trained deformable part-based models (DPM) \cite{DPM} and And-Or tree models \cite{xisong_cvpr}, modeling occlusion implicitly or explicitly \cite{boli_iccv13,VisualPhrases,siyu_bmvc,xiaogang_cvpr13,bojan_cvpr13}, and exploiting contextual information \cite{Ramanan_objpair,ramanan_layout,putting,autocontext,guangchen_cvpr}. The three aspects are usually studied separately.

\IEEEPARstart{C}{ar} is one of the most frequently seen object category in every day scenes. Car detection and viewpoint estimation by a computer vision system has broad applications such as autonomous driving and parking management. Fig.~\ref{fig:data} shows a few examples with varying complexities in car detection from four datasets. Car detection and viewpoint estimation are challenging problems due to the large structural and appearance variations, especially ubiquitous occlusions which further increase the intra-class variations significantly.
In this paper, we are interested in learning a unified model which can  detect cars in the four datasets and estimate car viewpoints. We aim to address two main issues in the following.  

\begin{figure} %[!ht]
\centering
%\framebox
{\includegraphics[width = 0.5\textwidth]{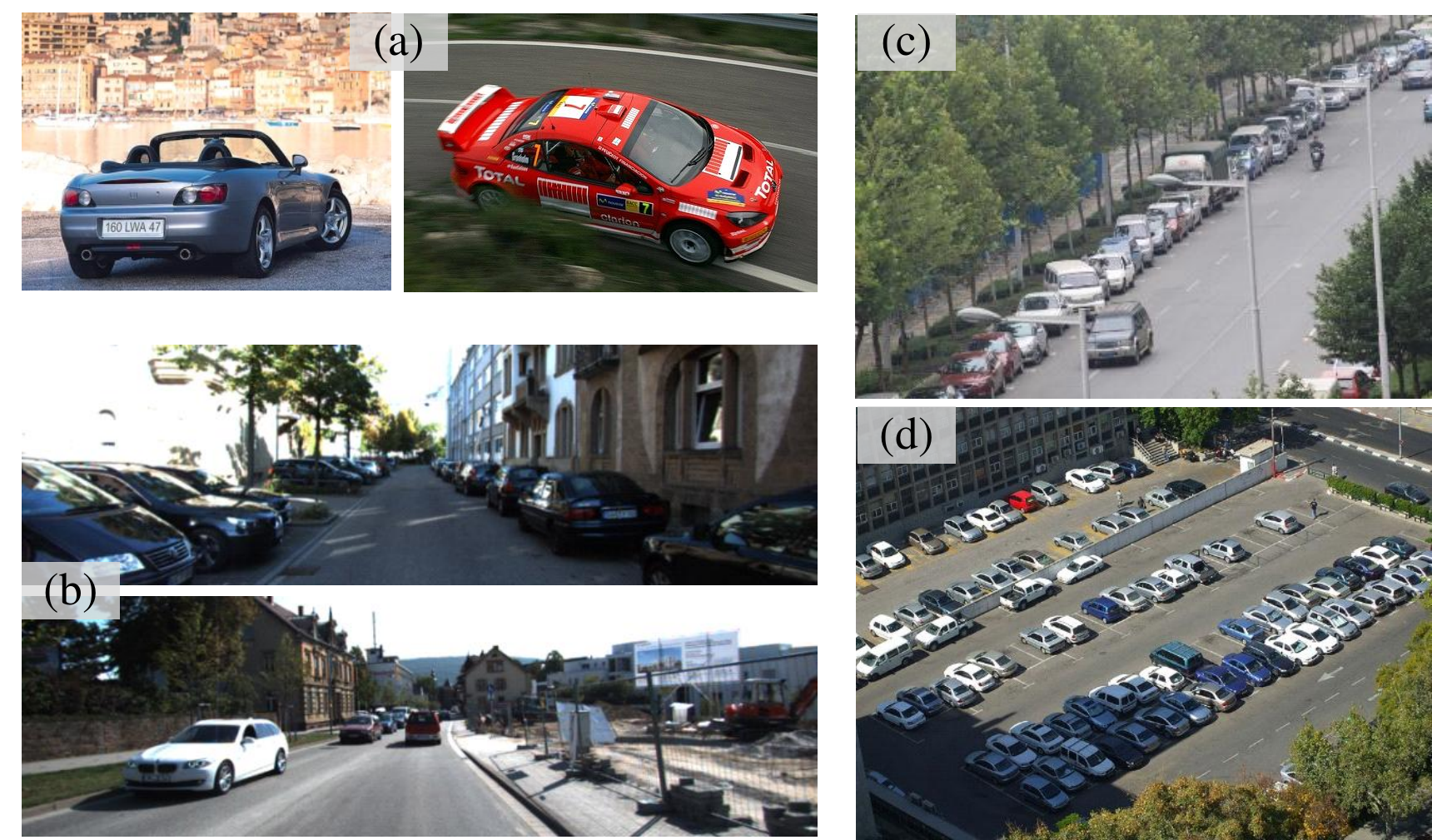}}
\caption{Illustration of varying complexities in car detection from four datasets. (a) The PASCAL VOC2007 car dataset~\cite{pascal} consists of single cars under different viewpoints but with less occlusion as pointed out in \cite{HoiemDiagnoseError}. (b) The KITTI car benchmark~\cite{Geiger12} includes on-road cars captured by a camera mounted upon a driving car which have more occlusions but restricted viewpoints. (c) The Street-Parking car dataset~\cite{boli_iccv13} includes cars with heavy occlusions but less multi-car context and (d) The Parking-Lot car dataset~\cite{boli_eccv14} consists of cars with heavy occlusions and rich multi-car context. The proposed And-Or model is learned for car detection in all four datasets.}
\label{fig:data}
\vspace{-4mm}
\end{figure} 

The first is to explicitly represent occlusion. Occlusion is a critical aspect in object detection for several reasons: (i) we do not know ahead of time what portion of an object (e.g. car) will be visible in a test image; (ii) we also do not know the occluded areas in weakly-labeled training data (i.e. only bounding boxes of single cars are given, as considered in this paper); and (iii) object occlusions in testing data could be very different from those in training data. 
Handling occlusions entails models capable of capturing the underlying regularities of occlusions at part level (i.e. different occlusion configurations). %It enforces part-based representation in general because dealing with occlusion entails searching through, or marginalizing, all possible occlusion configurations on-the-fly, and estimating viewpoints under occlusion entails information from visible parts.
%The recent literature of object detection has been focused on three aspects to improve accuracy performance: using hierarchical models such as discriminatively trained deformable part-based models (DPM) \cite{DPM} and And-Or tree models \cite{xisong_cvpr}, modeling occlusion implicitly or explicitly \cite{boli_iccv13,VisualPhrases,siyu_bmvc,xiaogang_cvpr13,bojan_cvpr13}, and exploiting contextual information \cite{Ramanan_objpair,ramanan_layout,putting,autocontext,guangchen_cvpr}. The three aspects are usually studied separately. 
\begin{figure*} %[!ht]
\centering
%\framebox
{\includegraphics[width = 1.0\textwidth]{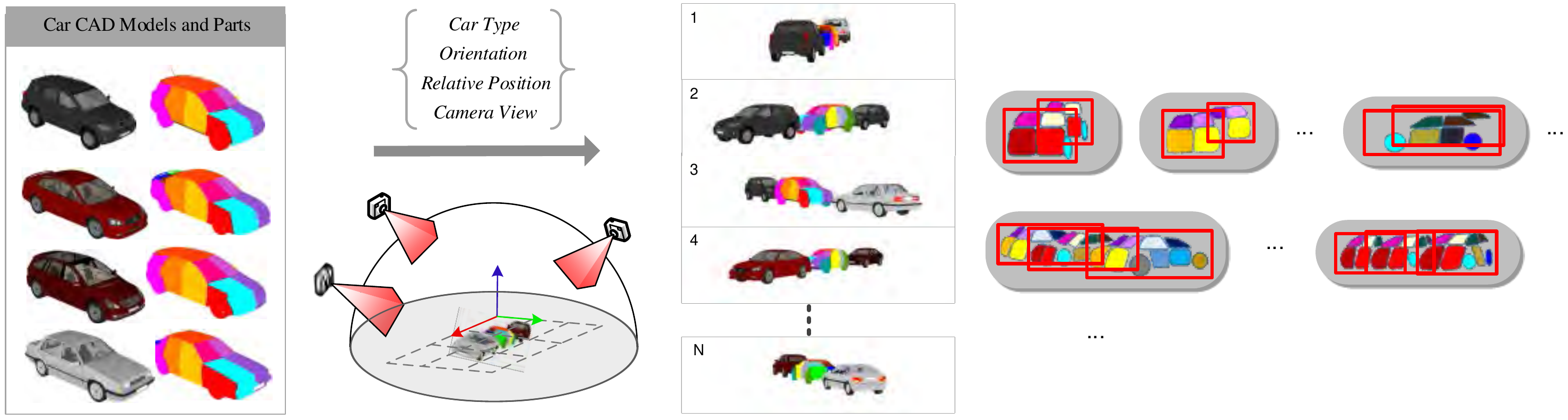}}
\caption{Illustration of the statistical regularities of car occlusions and multi-car contextual patterns by CAD simulation. We represent car-to-car occlusion at semantic part level (left) and generate a large number of synthetic occlusion configurations (middle) w.r.t. four factors (car type, orientation, relative position and camera view). We represent the regularities of different combinations of part visibilities (i.e., occlusion configurations) by a hierarchical And-Or model. This model also represents multi-car contextual patterns (right) based on the geometric configurations of single cars.}
\label{fig:regularity}
\vspace{-4mm}
\end{figure*} 

The second is to explicitly exploit contextual information co-occurring with occlusions (see examples in Fig.\ref{fig:data} (b), (c) and (d)), which goes beyond single-car detection. We focus on car-to-car contextual patterns (e.g., different multi-car configurations such as $2, 3$ or $4$ cars), which will be utilized in detection and viewpoint estimation and naturally integrated with occlusion configurations. 
 
 To represent both occlusion and context, we propose to learn an And-Or model which takes into account structural and appearance variations at multi-car, single-car and part levels jointly. Our And-Or model belongs to grammar models \cite{grammar,DPMGrammar} embedded in a hierarchical graph structure, which can express a large number of configurations (occlusion configurations and multi-car configurations) in a compositional and reconfigurable manner. Fig.\ref{fig:demo} illustrates our And-Or model.  
 By reconfigurable, it means that we learn appearance templates and deformation models for single cars and parts, and the composed appearance templates for a multi-car contextual pattern is inferred on-the-fly in detection according to the selections of their child single car Or-nodes. So, our model can express a large number of multi-car contextual patterns with different compatible occlusion configurations of single cars. \textit{Reconfigurability} is one of the most desired property in hierarchical models, which plays the main role in boosting the performance in our experiments, and also distinguishes the proposed method to other models such as the visual phrase model~\cite{VisualPhrases} and different object-pair models~\cite{ModelingOcclusion_PR2014,siyu_bmvc,xiaogang_cvpr13,bojan_cvpr13}.

%\vspace{-3mm}
\subsection{Method Overview}
\subsubsection{Data Preparation with Simulation Study} \label{sec:simulation}
Manually annotating car views, parts and part occlusions on real images are time-consuming and usually error-prone. One innovation in this paper is that we generate a large set of occlusion configurations and multi-car configurations by CAD models~\footnote{we used 40 CAD models selected from www.doschdesign.com and Google 3D warehouse} and a publicly available graphics rendering engine, the SketchUp SDK~\footnote{www.sketchup.com}. In the CAD simulation, the occlusion configurations and multi-car contextual patterns reflect variations in four factors: \textit{car type, orientation, relative position and camera view}. We decompose a car into $17$ semantic parts as shown  in different colors in the left side of Fig.~\ref{fig:regularity}. We then  generate a large number of examples by placing 3 cars in a $3\times 3$ grid (resembling the regularities of cars in parking lots or on the road, see the middle of Fig.~\ref{fig:regularity}). For the cars in the center, we compare their part visibilities from different viewpoints (as illustrated by the camera icons), and obtain \textit{the part occlusion data matrix} (each row represents an example and each entry takes a binary value, 0/1, representing occluded or not for a part under a viewpoint). The data matrix is used to learn the occlusion configurations. Similarly, we learn different multi-car contextual patterns based on the geometric configurations (see some examples in the right side of Fig.~\ref{fig:regularity}). Note that the semantic part annotations in the synthetic examples are used to learn the structure of our And-Or model and the parts are treated as latent variables in weakly-annotated training data of real images. We do not evaluate the performance of part localization and instead evaluate the viewpoint estimation based on the inferred part configurations. 

In the simulation, we place 3 cars in a $3\times 3$ grid with three considerations: (i) It can generate different occlusion configurations for the car in the center under different camera viewpoints, as well as different multi-car contextual patterns (2-car or 3-car pattern), which is easier than using 2 cars in processing the data in simulation. (ii) It can generate the synthetic dataset in which the occlusion configurations and multi-car contextual patterns are generic enough to cover the four situations in Fig.\ref{fig:data}. (iii) It can also reduce the gap between the synthetic data and real data when learning the initial appearance parameters for parts with the car in the back  instead of the white background (see more details in Sec.\ref{sec:learning}).        

\begin{figure*} %[!ht]
	\centering
	%\framebox
	{\includegraphics[width = 1.0\textwidth]{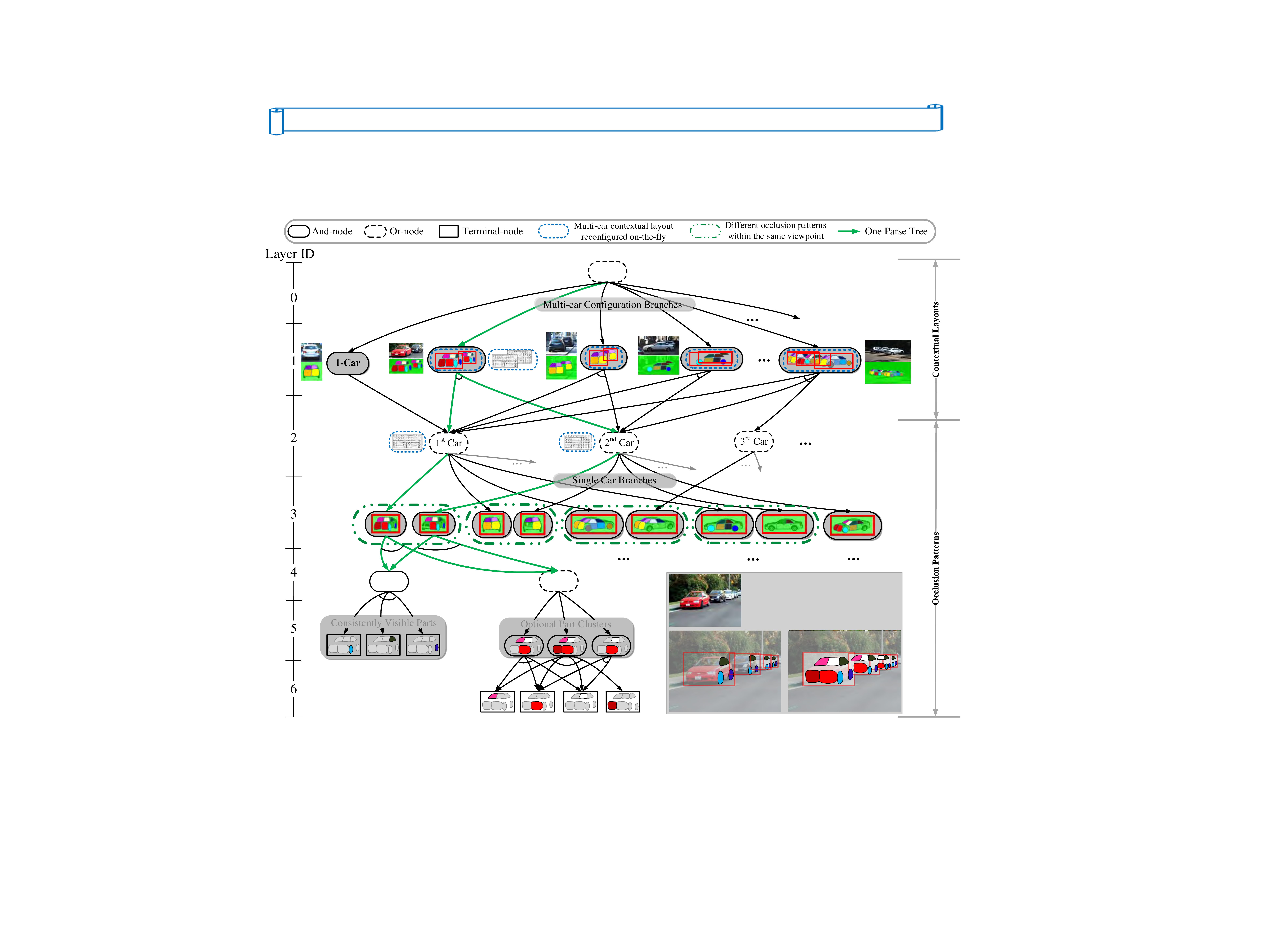}}
	\caption{Illustration of our And-Or model for car detection. It represents  multi-car contextual patterns and occlusion configurations jointly by modeling spatially-aligned multi-cars together and composing visible parts explicitly for single cars.  (Best viewed in color)}
	\label{fig:demo}
	\vspace{-3mm}
\end{figure*} 

%\vspace{-2mm}
\subsubsection{The And-Or Model}
There are three types of nodes in the And-Or model: an \textit{And-node} represents decomposition (e.g., a car is composed of a small number of parts), an \textit{Or-node} represents alternative ways of decomposition accounting for structural variations (e.g., different part configurations of a single car due to occlusions), and a \textit{Terminal-node} captures appearance variations to ground a car or a part to image data.
 
Fig. \ref{fig:demo} illustrates the learned And-Or model. The hierarchy consists of a layer of multi-car contextual patterns (top) and several layers of occlusion configurations of single cars (bottom).  The overall structure is as-follows: 

\textit{i) The root Or-node} represents different multi-car configurations which capture both viewpoints and car-to-car contextual patterns. Each multi-car contextual pattern is then represented by an And-node (e.g., car pairs and car triples shown in the figure). The contextual information  reflect the layout regularities of a small number, $N$  (e.g., $N\in \{2,3\}$), of cars in real sitations (such as cars in a parking lot).

\textit{ii) A multi-car And-node} is decomposed into nodes representing single cars. Each single car is represented by an Or-node (e.g., the $1^{st}$ car and the $2^{nd}$ car), since we have different combinations of car types, viewpoints and occlusion configurations.% (e.g., the car in the back of a car-pair can have different occluding situations due to the layouts). 
Here, a multi-car And-node embeds the reconfigurable compositional grammar of a multi-car configuration (e.g., the three 2-car configurations in the right-top of Fig.\ref{fig:regularity}) in which the single cars are reconfigurable w.r.t. viewpoint and occlusion configuration (up to some extend), and car type. This reconfigurability gives our model expressive power to handle the large variations of multi-car configurations in real sitations.   

\textit{iii) Each occlusion configuration is represented by an And-node} which is further decomposed into parts. Parts are learned using CAD  simulation (i.e., the $17$ semantic parts) and are organized into consistently visible parts and optional part clusters (see the example in the right-bottom of Fig.~\ref{fig:demo}). Then, a single car can be represented by the consistently visible parts (i.e., And) and one of the
optional part clusters (i.e., Or).
The green dashed bounding boxes show some examples corresponding to different occlusion configurations (i.e., visible parts) from the same viewpoint.

%\vspace{-1mm}
\subsubsection{Weakly-supervised Learning of the And-Or Model}
Using weakly-annotated real image training data and the synthetic data, we learn the And-Or model in two stages: 

\textit{i) Learning the structure of the hierarchical And-Or model}. Both the multi-car contextual patterns and occlusion configurations of single cars are learned automatically based on the annotated single car bounding boxes in training data together with the synthetic examples generated from CAD simulations. The multi-car contextual patterns are mined or clustered from the geometric layout features. The occlusion configurations are learned by a clustering method using the part visibility data matrix. The learned structure is a directed and acyclic graph since we have both single-car-sharing and part-sharing, thus Dynamic Programming (DP) can be applied in inference.

\textit{ii) Learning the parameters for appearance, deformation and bias}. Given the learned structure of the And-Or model, we jointly train the parameters in the structural SVM framework and adopt the Weak-Label Structural SVM (WLSSVM) method  \cite{pffgrammar,McAllesterLossBound} in implementation.

%The proposed And-Or model is flexible and reconfigurable to account for the large variations of both multi-car context and occlusion configurations in complex situations. It represents occlusions at semantic part level and captures the regularities of different combinations of part visibilities (i.e., occlusion configurations). By reconfigurable, it means that we learn appearance templates and deformation models for single cars and parts, and the composed appearance templates for a multi-car contextual patter is inferred on-the-fly in detection according to the selections of their child single car Or-nodes. So, our model can express a large number of multi-car contextual patterns with different compatible occlusion configurations of single cars. \textit{Reconfigurability} is one of the most desired property in hierarchical models, which plays the main role in boosting the performance in our experiments, and also distinguishes the proposed method to other models such as the visual phrase model~\cite{VisualPhrases} and different object pair models~\cite{ModelingOcclusion_PR2014,siyu_bmvc,xiaogang_cvpr13,bojan_cvpr13}.

%\vspace{-2mm}
\subsubsection{Experiments}
In experiments, we evaluate the detection performance of our model on four car datasets: the KITTI dataset \cite{Geiger12}, the PASCAL VOC2007 car dataset \cite{pascal} and two self-collected datasets \--- the Street-Parking dataset \cite{boli_iccv13} and the Parking Lot dataset~\cite{boli_eccv14} (which are released with this paper). Our model outperforms different state-of-the-art variants of DPM \cite{DPM} (including the latest implementation \cite{voc5}) on all the four datasets, as well as other state-of-the-art models \cite{behley2013iros,Geiger11,bojan_cvpr13,boli_iccv13} on the KITTI and the Street-Parking datasets. We evaluate viewpoint estimation performance on three car datasets: the PASCAL VOC2006 car dataset~\cite{pascal}, the 3D car dataset~\cite{savarese}, and the PASCAL3D+ car dataset~\cite{xiang_wacv14}. Our model achieves comparable performance with the state-of-the-art methods (significantly better than the method using deep learning features~\cite{pedersoli}). 
\textit{The detection code and data are available on the author's homepage} \footnote{http://www.stat.ucla.edu/\textasciitilde tfwu/projects.htm}. 

\textbf{Paper Organization.} The remaining of this paper is organized as follows. Section \ref{sec:relatedwork} overviews the related work and summarizes our contributions. Section \ref{sec:model} presents the And-Or model and defines its scoring functions. Section \ref{sec:mining} presents the method of mining  multi-car contextual patterns and occlusion configurations of single cars in weakly-labeled training data. Section \ref{sec:learning} discusses the learning of model parameters using WLSSVM, as well as details of the DP inference algorithm. Section \ref{sec:exp} presents the experimental results and comparisons of the proposed model on the four car detection datasets and the three viewpoint estimation datasets. Section \ref{sec:conclusion} concludes the paper with discussions. 

%-------------------------------------------------------------------------
%\vspace{-2mm}
\section{Related Work and Our Contributions}\label{sec:relatedwork} %\vspace{-2mm}
Over the last decade, object detection has made much progress in various vision tasks such as face detection \cite{vj_detector}, pedestrian detection \cite{Dollar2012PAMI}, and generic object detection \cite{pascal,DPM,xisong_cvpr}. In this section we focus on occlusion and context modeling in object detection, and classify the recent literature into three research streams. For a full review of contemporary approaches, we refer the reader to recent survey articles \cite{recog_Grauman,survey_50,survey_zhang}.

\textit{i) Single Object Modeling and Occlusion Modeling.} Hierarchical models are widely used in the recent literature of object detection and most existing approaches are devoted to learning a single object model. Many work extended the deformable part-based model \cite{DPM} (which has a two-layer structure) by exploring deeper hierarchy and global part configurations~\cite{xisong_cvpr,leo2010,pffgrammar}, using strong manually-annotated parts \cite{StrongSupervsionDPM} or CAD models \cite{teach3D}, or keeping human in-the-loop \cite{interactiveDPM}. 
To address the occlusion problem, various occlusion models estimate the visibilities of parts from image appearance, using assumptions that the visibility of a part is (a) independent from other parts \cite{wubo,hoglbp,HR_car,landmark_iccv13,desai_eccv12}, (b) consistent with neighboring parts \cite{Gao2011,pffgrammar}, or (c) consistent with its parent or child parts describing object appearance at different scales \cite{duan2010}. 
Another essential problem is to organize part configurations. Recently, \cite{pffgrammar, HR_car, boli_iccv13} explored different ways to deal with this problem. In particular, \cite{HR_car} modeled different part configurations by the local part mixtures. \cite{pffgrammar} used a more flexible grammar model to infer both the occluder and visible parts of an occluded person. \cite{boli_iccv13} regularized parts into consistently visible parts and optional part clusters, which is more efficient to represent occlusion configurations. 
Recent work~\cite{Mathias2013Iccv,consensus,zia,Ghiasi_face14,Ghiasi_people14} proposed to enumerate possible occlusion configurations and model each occlusion configuration as a specific component.
{\cite{OhnBar} proposed a 2D model to learn discriminative subcategories, and  \cite{xiang_cvpr15} further integrated it with an explicit 3D occlusion model, both showing excellent performance on the KITTI dataset.}
Though those models were successful in some heavily occluded cases, they did not represent contextual information, and usually learned  another separate context model using the detection scores as input features. Recently, an And-Or quantization method was proposed to learn And-Or tree models \cite{xisong_cvpr,Tangram} for generic object detection in PASCAL VOC~\cite{pascal} and learn 3D And-Or models \cite{Wenze3DCar} respectively, which could be useful in occlusion modeling.  

\textit{ii) Object-Pair and Visual Phrase Models}. To account for the strong co-occurrence, object-pair~\cite{ModelingOcclusion_PR2014,siyu_bmvc,xiaogang_cvpr13,bojan_cvpr13} and visual phrase~ \cite{VisualPhrases} methods modeled occlusions and interactions using a X-to-X or X-to-Y composite template that spans both one object (i.e., ``X" such as a person or a car) and another interacting object (i.e., ``X" or ``Y" such as the other car in a car-pair in parking lots or a bicycle on which a person is riding).  Although these models can handle occlusion better than single object models, the object-pair or visual phrase modeled occlusion implicitly, and they were often manually designed with fixed structures (i.e., not reconfigurable in inference). They performed worse than original DPM in the KITTI dataset as evaluated by  \cite{bojan_cvpr13}.

%\begin{figure*}%[t]
%\centering
%\framebox{\includegraphics[width = 0.85\textwidth]{./model.pdf}}
%\caption{The learned And-Or model for car detection (only a portion of the whole model  is shown here for clarity). The node in layer $0$ is the root Or-node, which has a set of child And-nodes representing different multi-car configurations in layer $1$ (the number of single cars $N\leq 2$ is considered). The nodes in layer $2$ represent single car Or-nodes, each of which has a set of child And-nodes representing single cars with different viewpoints and occlusion configurations. We learn appearance templates for single cars and their parts (nodes in layer $3$ and $4$), and the composite templates for a multi-car is reconfigured on-the-fly in inference (as illustrated by the green solid arrows).}
%\label{fig:model} %\vspace{-5mm}
%\vspace{-3mm}
%\end{figure*}

%\textit{Context Models.} Many context models have been exploited in object detection showing performance improvement~\cite{Ramanan_objpair,ramanan_layout,putting,autocontext,guangchen_cvpr}. In \cite{autocontext}, Tu and Bai integrate the detector responses with background pixels to determine the foreground pixels. In \cite{guangchen_cvpr}, Chen, et. al. propose a multi-order context representation to take advantage of the co-occurrence of different objects. Most of them model objects and context separately.

\textit{iii) Context Models.} Many context models have been exploited in object detection with improved performance~\cite{Ramanan_objpair,ramanan_layout,putting,autocontext,guangchen_cvpr}. Hoiem et al. \cite{putting} explored a scene context, Desai et al. \cite{ramanan_layout} improved object detectors by incorporating the multi-class context on the pascal dataset \cite{pascal} in a max-margin framework.
In \cite{autocontext}, Tu and Bai integrated the detector responses with background pixels to determine the foreground pixels. In \cite{guangchen_cvpr}, Chen et. al. proposed a multi-order context representation to take advantage of the co-occurrence of different objects. 
Recently, \cite{nyc3dcar} explored geographic contextual information to facilitate  car detection, and \cite{pano} explored a 3D panoramic context in object detection.
Although these work verified that context is crucial in object detection, most of them modeled objects and context separately, not in a unified framework.

This paper is extended from our two previous conference papers \cite{boli_eccv14,boli_iccv13} in the following aspects: (i) A unified representation is learned for integrating occlusion and context; (ii) More details on the learning algorithm and the detection algorithm are presented; (iii) More analyses and comparisons on the experimental results are added with improved performance. 

This paper makes three contributions to the literature of car detection. 

i) It proposes an And-Or model to represent multi-car context and occlusion configurations. The proposed model is multi-scale and reconfigurable to account for large structure, viewpoint and occlusion variations.

ii) It presents a simple, yet effective, approach to mine context and occlusion configurations from weakly-labeled training data.

iii) It introduces two datasets for evaluating occlusion and multi-car context, and obtains performance comparable to or better than state-of-the-art car detection methods in four challenging datasets.

%\vspace{-4mm}
%------------------------------------------------------------------------
\section{Representation and Inference} \label{sec:model} 
\subsection{The And-Or Model and Scoring Functions}
In this section, we introduce the notations in defining the And-Or model and its scoring functions. 

An \textit{And-Or model} is defined by a $3$-tuple, 
$
\mathcal{G} = (\mathcal{V}, E, \Theta),
$
where 
$
\mathcal{V} = \mathcal{V}_\AND \cup \mathcal{V}_\OR \cup \mathcal{V}_T,
$ 
represents the nodes in three subsets: And-nodes $\mathcal{V}_\AND$, Or-nodes $\mathcal{V}_\OR$ and Terminal-nodes $\mathcal{V}_T$; $E$ is the set of edges organizing all the nodes in a directed and acyclic graph (DAG);    
$
\Theta=(\Theta^{app}, \Theta^{def}, \Theta^{bias}),
$ 
is the set of parameters (for appearance, deformation and bias respectively, to be defined later). %Fig.～\ref{fig:demo} shows the learned car And-Or model which has five layers. 

A \textit{Parse Tree} is an instantiation of the And-Or model by selecting the best child (according to the scoring functions to be defined) for each encountered Or-node. The green arrows in Fig.~\ref{fig:demo} show an example of parse tree.

\textit{Appearance Features.} We adopt the Histogram of Oriented Gradients (HOG) feature \cite{HOG,DPM} to describe appearance. Let $I$ be an image defined on an image lattice. Denote by $\mathcal{H}$ the HOG feature pyramid computed for $I$ using $\lambda$ levels per octave, and by $\Lambda$ the lattice of the whole pyramid. Let $p=(l, x, y)\in \Lambda$ specify a position $(x, y)$ in the $l$-th level of the pyramid $\mathcal{H}$. Denote by $\Phi^{app}(\mathcal{H}, p_t)$ the extracted HOG features for a Terminal-node $t$ placing at position $p_t$ in the pyramid.

\textit{Deformation Features.} We allow local deformation when composing the child nodes into a parent node. In our model, parts are placed at twice the spatial resolution w.r.t. single cars, while single cars and composite multi-cars are at the same spatial resolution. We penalize the displacements between the anchor locations of child nodes (w.r.t. the placed parent node) and their actual deformed locations.
 Denote by $\delta=[dx, dy]$ the displacement. The deformation feature is defined by,
 \[
 \Phi^{def}(\delta) = [dx^2, dx, dy^2, dy]'.
 \]
%Denote by $\theta_{v|u}^{geo}$ is the geometry parameter for node $X$ w.r.t $Y$, and $\Phi^{geo}_{X|Y}(q,p)$ is the corresponding geometry feature, which is same as the quadratic penalty in DPM. Thus $\Phi^{geo}_{X|Y}(q,p) = [\mathrm{d}u^2, \mathrm{d}v^2, \mathrm{d}u,\mathrm{d}v]$, where $\mathrm{d}u$ and $\mathrm{d}v$ are the displacements of $q$ from $p$.

A \textbf{Terminal-node} $t\in \mathcal{V}_T$ grounds a single car or a part to image data (see Layer 3 and 4 in Fig.\ref{fig:demo}). Given a parent node $A$, the model for $t$ is defined by a 4-tuple 
\[(\theta_t^{app}, s_t, a_{t|A}, \theta^{def}_{t|A})
\] 
where $\theta_t^{app}\subset \Theta^{app}$ is the appearance template, $s_t\in\{0, 1\}$ the scale factor for placing node $t$ w.r.t. its parent node, $a_{t|A}$ a two-dimensional vector specifying an anchor position relative to the position of parent node $A$, and $\theta^{def}_{t|A}\subset \Theta^{def}$ the deformation parameters. Given the position $p_A=(l_A, x_A, y_A)$ of the parent node $A$, the scoring function of a Terminal-node $t$ is defined by,
\begin{align}
\nonumber score(t|A, p_A) = \max_{\delta \in \Delta}(&<\theta_t^{app}, \Phi^{app}(\mathcal{H}, p_t)> -\\
& <\theta^{def}_{t|A}, \Phi^{def}(\delta)>), \label{eq:tscore}
\end{align}
where $\Delta$ is the space of deformation (i.e., the lattice of the corresponding level in the feature pyramid), $p_t=(l_t, x_t, y_t)$ with  $l_t=l_A-s_t\lambda$ and $(x_t, y_t)=2^{s_t}(x_A, y_A)+a_{t|A}+\delta$ where $s_t=0$ means the object and parts are placed at the same resolution and $s_t=1$ means parts are placed at twice the resolution of the object templates, and $<\cdot, \cdot>$ denotes the inner product. Fig.\ref{fig:demo} shows some learned appearance templates. 

An \textbf{And-node} $A\in \mathcal{V}_{\AND}$ represents a decomposition of a large entity (e.g., a multi-car layout at Layer 1 or a single car at Layer 3 in Fig.\ref{fig:demo}) into its constituents (e.g., $2$ or $3$ single cars or a small number of  parts). Single car And-nodes are associated with viewpoints. Unlike the Terminal-nodes, single car And-nodes are not allowed to be deformable in a multi-car configuration in this paper (we implemented it in experiments and did not observe performance improvement, so for simplicity we make them not deformable).  Denote by $ch(v)$ the set of child nodes of a node $v\in \mathcal{V}_\AND \cup \mathcal{V}_\OR$. The position $p_A$ of an And-node $A$ is inherited from its parent Or-node, and then the scoring function is defined by, 
\begin{equation}
score(A, p_A) = \sum_{v\in ch(A)} score(v|A, p_A) + b_A \label{eq:andscore}
\end{equation}
where $b_A\in \Theta^{bias}$ is the bias term. Each single car And-node (at Layer 3) can be treated as the DPM \cite{DPM} or the And-Or structure proposed in \cite{boli_iccv13}. So, our model is flexible to  integrate state-of-the-art single object models. For multi-car And-nodes (at Layer 1), their child nodes are Or-nodes and the scoring function $score(v|A, p_A)$ is defined below. 

An \textbf{Or-node} $O\in \mathcal{V}_\OR$ represents different structure variations (e.g., the root node and the $i$-th car node at Layer 2 in Fig.\ref{fig:demo}). For the root Or-node $O$, when placing at the position $p\in \Lambda$, the scoring function is defined by,
\begin{equation}
score(O, p) = \max_{v\in ch(O)} score(v, p), \label{eq:orscore1}
\end{equation}
where $ch(O)\subset \mathcal{V}_\AND$. For the $i$-th car Or-node $O$, given a parent multi-car And-node $A$ placed at $p_A$, the scoring function is then defined by,
\begin{align}
\nonumber score(O|A, p_A) = \max_{v\in ch(O)} \max_{\delta\in \Delta}(&score(v, p_v) -\\ &<\theta_{O|A}^{def}, \Phi^{def}(\delta)>), \label{eq:orscore2}
\end{align}
where $p_v=(l_v, x_v, y_v)$ with $l_v = l_A$ and $(x_v, y_v) = (x_A, y_A) + \delta$. The best child of an Or-node is computed by taking $\argmax$ of Eqn.(\ref{eq:orscore1}) and Eqn.(\ref{eq:orscore2}).

%\vspace{-2mm}
\subsection{The DP Algorithm in Detection}
In detection, we place the And-Or model at all positions $p\in \Lambda$ and retrieve the optimal parse trees for all positions at which the scores are greater than the detection threshold. 
Thank to the directed and acyclic structure of our And-Or model, we can utilize the efficient DP algorithm which consists of two stages:

\textit{In the bottom-up pass:} Following the depth-first-search (DFS) order of nodes in the And-Or model, the bottom-up pass computes the matching scores of all possible parse trees of the And-Or model at all possible positions in the whole feature pyramid.  %appearance score maps and deformed score maps for the whole feature pyramid $\mathcal{H}$ for all Terminal-nodes, And-nodes and Or-nodes. The deformed score maps can be computed efficiently by the generalized distance transform algorithm as done in \cite{DPM}.

First of all, we compute the appearance score maps (pyramid) for all Terminal-nodes (which is done by filter convolution). The optimal position of a Terminal-node w.r.t. a parent node can be computed as a function of the position of the parent node. The quality (matching score) of the optimal position for a Terminal-node w.r.t. a given position of the parent is computed using Eqn.\ref{eq:tscore} (which yields the deformed score map through the generalized distance transform trick as done in the DPM \cite{DPM} for efficiency), and the optimal position can be retrieved by replacing $\max$ in Eqn.(\ref{eq:tscore}) with $\arg \max$.

Then, following the DFS order of nodes, we compute the score maps for all the And-nodes and Or-nodes using Eqn.(\ref{eq:andscore}), (\ref{eq:orscore1}) and (\ref{eq:orscore2}) with the score maps of their child nodes having been computed already. Similarly, we can obtain the optimal branch for each Or-node by replacing the $\max$ in Eqn.(\ref{eq:orscore1}) and (\ref{eq:orscore2}) with $\arg \max$. 

\textit{In the top-down pass}, we first find all detection candidates for the root Or-node $O$ based on its score maps, i.e., the positions 
$
\mathbb{P} = \{p; score(O,p) \geq \tau \text{ and } p \in \Lambda\}.
$
Then, following the breadth-first-search (BFS) order of nodes, we retrieve the optimal parse tree at each $p\in \mathbb{P}$: starting from the root Or-node, we select the optimal branch of each encountered Or-node, keep all the child nodes of each encountered And-node, and retrieve the optimal position of each Terminal-node. Based on the parsed sub-tree rooted at single car And-nodes, we obtain the viewpoint estimation and the occlusion configuration. 

\textit{Post-processing.}
To generate the final detection results of single cars for evaluation, we apply multi-car guided non-maximum suppression (NMS) to deal with occlusions: 

i) Some of the single cars in a multi-car detection candidate are highly overlapped due to occlusion, so if we directly use conventional NMS, we will miss the detection of the occluded cars. We enforce that all the single car bounding boxes in a multi-car prediction will not be suppressed by each other. A similar idea is also used in \cite{siyu_bmvc}. 
	
ii) Overlapped multi-car detection candidates might report multiple predictions for the same single car. For example, if a car is shared by a $2$-car detection candidate and a $3$-car detection candidate, it will be reported twice. We will keep only the one with higher score.

%------------------------------------------------------------------------
%\vspace{-2mm}
\section{Learning And-Or Structures} \label{sec:mining} %\vspace{-2mm}
In this section, we present the methods of learning the structures of And-Or model by mining contextual patterns and occlusion configurations in the positive training dataset. 

%\vspace{-2mm}
\subsection{Generating Multi-car Training Samples}
\textbf{Positive Samples.} Denote by $D^+=\{(I_1, \mathbb{B}_1), \cdots, (I_n, \mathbb{B}_n)\}$ the positive training dataset with  $\mathbb{B}_i = \{B_i^j = (x_i^j, y_i^j, w_i^j, h_i^j)\}_{j=1}^{k_i}$ being the set of $k_i$ annotated single car bound boxes in image $I_i$. Here, $(x, y)$ is the left-top corner and $(w, h)$ the width and height. 

Denote the set of $N$-car positive samples by,
\begin{equation}
D_{N\mbox{-}car}^+ = \{(I_i, B_i^J); {|J| = N, B_i^J\subseteq \mathbb{B}_i, i\in [1, n]} \}. 
\end{equation}
where all the $I_i$'s have more than $N$ annotated single cars (i.e., $k_i\geq N$).  
We have, 

i) $D_{1-car}^+$ consists of all the single car bounding boxes which do not overlap the other ones in the same image. For $N\geq 2$, $D_{N-car}^+$ is generated iteratively. 

ii) In generating $D_{2-car}^+$ (see Fig.\ref{fig:trainingData} (a)), for each positive image $(I_i, \mathbb{B}_i)\in D^+$ with $k_i\geq 2$, we enumerate all valid $2$-car configurations starting from $B_i^1\in \mathbb{B}_i$: we first select the current $B_i^j$ as the first car ($1\leq j \leq k_i$), obtain all the surrounding car bounding boxes  $\mathcal{N}_{B_i^j}$ which overlap $B_i^j$, and then select the second car $B_i^{k}\in \mathcal{N}_{B_i^j}$ which has the largest overlap if $\mathcal{N}_{B_i^j}\neq \emptyset$ and $(I_i, B_i^J) \notin D^+_{2-car}$ ($J=\{j,k\}$).

iii) In generating $D_{N-car}^+$ ($N>2$, see Fig.\ref{fig:trainingData} (b)), for each positive image with $k_i\geq N$ and $\exists (I_i, B_i^K)\in D_{(N-1)-car}^+$, we first select the current $B_i^K$ as the seed, obtain the neighbors $\mathcal{N}_{B_i^K}$ each of which overlaps at least one bounding box in $B_i^K$, and then select the bounding box $B_i^j\in \mathcal{N}_{B_i^K}$ which has the largest overlap and add $(I_i, B_i^J)$ to $D_{N-car}^+$ ($J = K \cup \{j\}$).

\begin{figure} %[!ht]
	\centering
	%\framebox
	{\includegraphics[width = 0.46\textwidth]{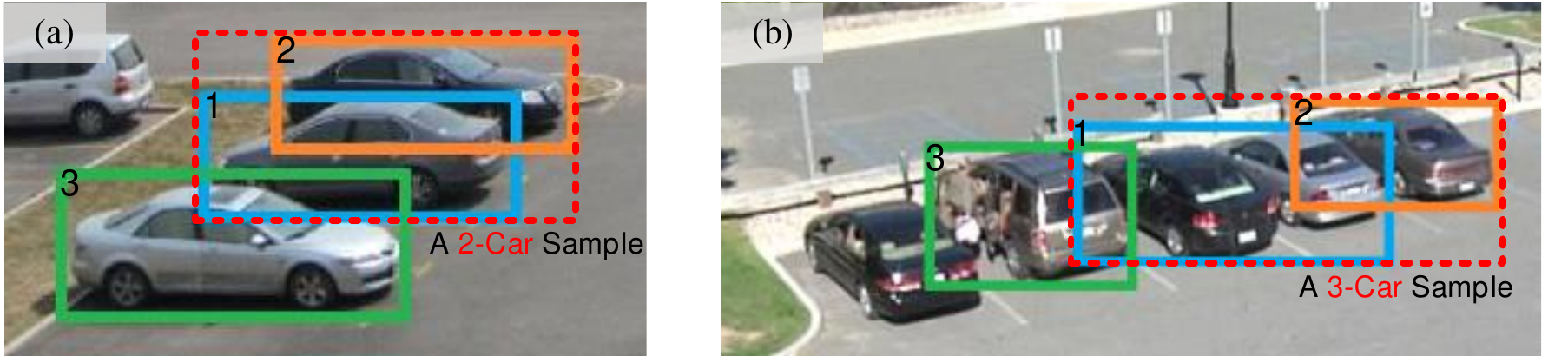}}
	\caption{Illustration of generating multi-car positive samples.}
	\label{fig:trainingData}
	\vspace{-4mm}
\end{figure} 

\textbf{Negative Samples.} We collect negative samples in images without cars appearing provided in the benchmark datasets and apply the hard negative mining approach during learning parameters as done in the DPM \cite{DPM}. 

%\begin{figure*}[!t]
%\centering
%%\framebox
%{\includegraphics[width = 0.75\textwidth]{./context.pdf}}
%\caption{
%\textit{Top}: $2$-car context patterns on the KITTI dataset \cite{Geiger12} and self-collected Parking Lot dataset. Each context pattern is represented by a specific color set, and each circle stands for the center of each cluster.
%\textit{Middle}: Overlap ratio histograms of the KITTI dataset and the Parking Lot dataset (we show the occluded cases only). 
%\textit{Bottom}: some cropped examples with different occlusions. The $2$ bounding boxes in a car pair are shown in red and blue respectively. (Best viewed in color).}
%\label{fig:hist} 
%\vspace{-3mm}
%\end{figure*}

\begin{SCfigure*}
  \centering
  \caption{\textit{Left-Top}: $2$-car context patterns on the KITTI dataset \cite{Geiger12} and self-collected Parking Lot dataset. Each context pattern is represented by a specific color set, and each circle stands for the center of each cluster.
  \textit{Left-Bottom}: Overlap ratio histograms of the KITTI dataset and the Parking Lot dataset (we show the occluded cases only). 
  \textit{Right}: some cropped examples with different occlusions. The $2$ bounding boxes in a car pair are shown in red and blue respectively. (Best viewed in color).}\label{fig:hist} \vspace{-3mm}
  \includegraphics[width=0.78\textwidth]%
    {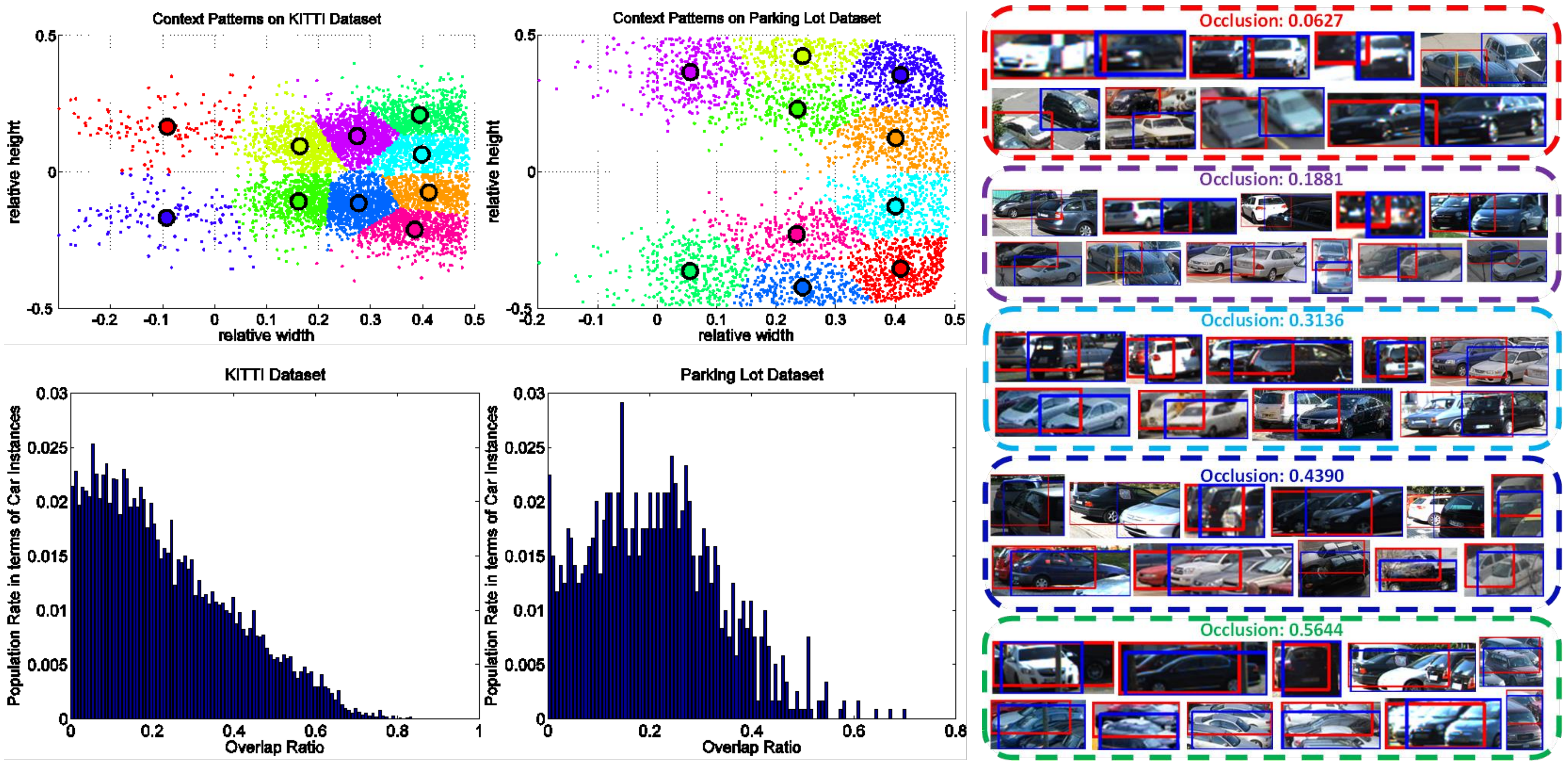}% picture filename
\end{SCfigure*}

%\vspace{-2mm}
\subsection{Mining Multi-car Contextual Patterns} \label{sec:cm} %\vspace{-2mm}
This section presents the method of learning multi-car patterns in Layer $0-2$ in Fig.\ref{fig:demo}. 
 Considering $N\geq 2$, we use the relative positions of single cars to describe the layout of a multi-car sample $(I_i, B_i^J)\in D^+_{N-car}$. Denote by $(cx, cy)$ the center of a car bounding box ($J=\{1, \cdots, N\}$). Let $w_{J}$ and $h_{J}$ be the width and height of the union bounding box of $B_i^J$ respectively. With the center of the first car being the centroid, we define the layout feature by,
 \begin{equation}
 [{{cx_i^2-cx_i^1}\over w_{J}}, {{cy_i^2-cy_i^1}\over h_{J}}, \cdots, {{cx_i^N-cx_i^1}\over w_{J}}, {{cy_i^N-cy_i^1}\over h_{J}}].
 \end{equation}
 
  We cluster these layout features over $D^+_{N-car}$ to get $T$ clusters using $k$-means. \textit{The obtained clusters are used to specify the And-nodes at Layer 1} in Fig.\ref{fig:demo}. The number of cluster $T$ is specified empirically for different training datasets in our experiments.

In Fig.~\ref{fig:hist} (top), we visualize the clustering results for $D_{2-car}^+$ on the KITTI \cite{Geiger12} and the Parking Lot datasets. 
Each set of color points represents a $2$-car context pattern. 
In the KITTI dataset, we can observe there are some car-to-car ``peak" modes in the dataset (similar to the analyses in \cite{bojan_cvpr13}), while the context patterns are more diverse in the Parking Lot dataset.

%\vspace{-2mm}
\subsection{Mining Occlusion Configurations} \label{sec:om} %\vspace{-2mm}
 In this section we present the method of learning occlusion configurations for single cars in Layer 3 and 4 in Fig.\ref{fig:demo}. 
We learn the occlusion configurations automatically from a large number of occlusion configurations generated by CAD simulations. Note that the synthetic data are used to learn the occlusion configurations, while the appearance and geometry parameters are still learned from real data. 

%\vspace{-2mm}
\subsubsection{Generating Occlusion Configurations}\label{sec::sampling}
As mentioned in Sec.\ref{sec:simulation},  we choose to put $3$ cars in generating  occlusion configurations. Specifically, we choose the center and 2 other randomly selected positions on a $3 \times 3$ grid, and put cars around these grid points to simulate occlusions. 
%For each set of the position triplet, we randomly choose values for a few factors controlling the occlusion, and then extract an occlusion configuration from the generated image. 
See some examples in Fig.\ref{fig:regularity}.

The occlusion configurations reflect the four factors: \emph{car type} $t$, \emph{orientation} $\rho$ , \emph {relative position} $r$ and \emph{camera view} $\Pi$. To generate an occlusion configuration, we randomly assign values for these factors, where for each car with type $i$, $\rho_i \in \{\textrm{frontal,rear}\}$, $r_i=r_i^{(0)} + \mathrm{d}r$, where $r_i^{(0)}$ is the nominated position for the $i$-$th$ car on the $3 \times 3$ grid, and $\mathrm{d}r = (\mathrm{d}x,\mathrm{d}y)$ is the relative distance (along $x$ axis and $y$ axis) between sampled position and nominated position of the $i$-$th$ car. The camera view is in the range of $\mathrm{azimuth} \in [0,2\pi]$ and $\mathrm{elevation} \in [0,\pi/4]$, we discretize the view space into $B$ view bins uniformly along the azimuth angle. 
In the synthesized configurations, a part is treated as occluded if $60\%$ of its area is not visible.

\begin{figure*} %[!ht]
\centering
%\framebox
{\includegraphics[width = 1.0\textwidth]{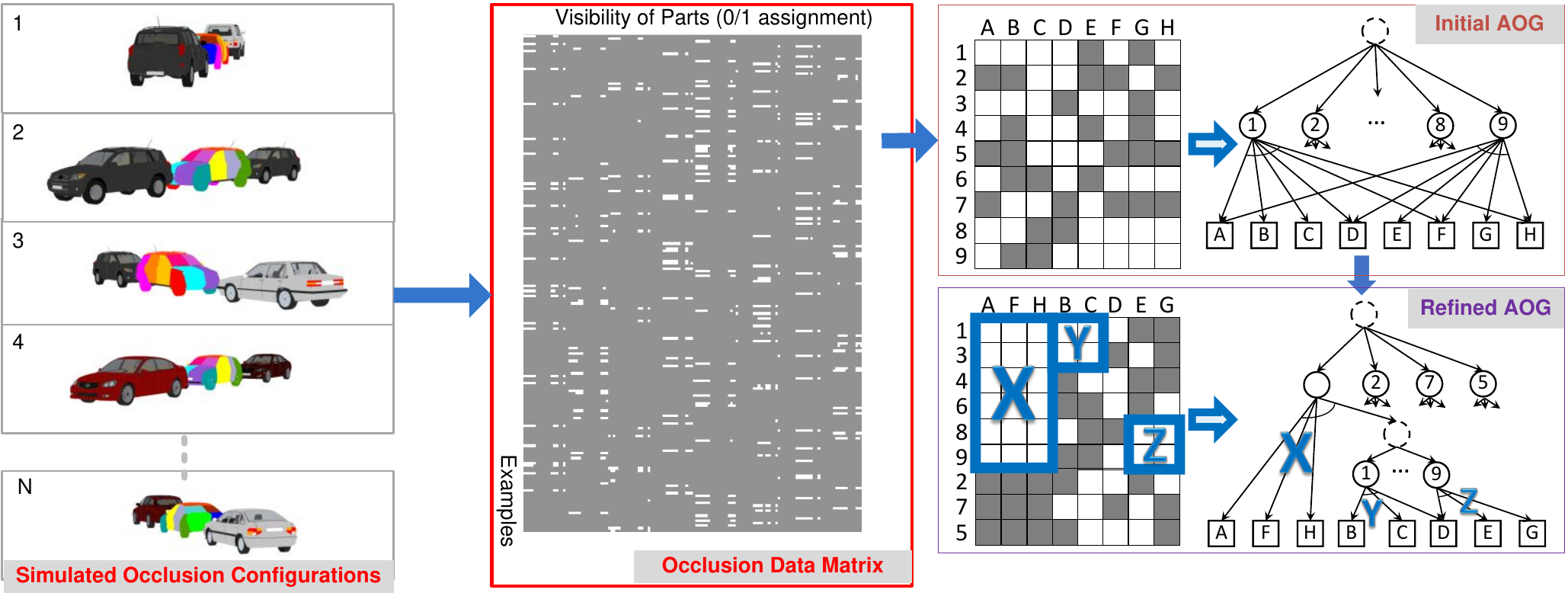}}
\caption{Illustration of learning occlusion configurations. It consists of three components: (i) Generating occlusion configurations using CAD simulations with 17 semantic parts in total; (ii) Learning the initial And-Or structure based on the data matrix constructed from the simulated occlusion configurations. Each row of the data matrix represents an example and the columns represent the visibility of the 17 semantic parts (a white/gray entry denotes a part is visible/invisible. Each example is represented by an And-node as one child of the root Or-node; (iii) Refining the initial And-Or structure using graph compression algorithm~\cite{zzAoT} to seek the consistently visible parts (e.g., $X$) and optional part clusters (e.g., $Y$ and $Z$).}
\label{fig:occlusion}
\vspace{-4mm}
\end{figure*} 

%\begin{SCfigure*}
%  \centering
%  \caption{Illustration of learning occlusion configurations. It consists of three components: (i) Generating occlusion configurations using car CAD models with 17 semantic parts in total; (ii) Learning the initial And-Or structure based on the data matrix constructed from the simulated occlusion configurations. Each row of the data matrix represents an example and the columns represent the visibility of the 17 semantic parts (a white/gray entry denotes a part is visible/invisible. Each example is represented by an And-node as one child of the root Or-node; (iii) Refining the initial And-Or structure using graph compression algorithm~\cite{zzAoT} to seek the consistently visible parts (e.g., $X$) and optional part clusters (e.g., $Y$ and $Z$).}
%  \label{fig:occlusion}\vspace{-3mm}
%  \includegraphics[width=0.7\textwidth]%
%    {./occlusion.pdf}% picture filename
%\end{SCfigure*}

%\vspace{-2mm}
\subsubsection{Constructing the Initial And-Or model of Single Cars}
With the part-level visibility information, we compute two vectors for each occlusion configuration: The first is a ($17$ parts$\times B$ camera views) dimension binary valued vector $\vec{v}$ for the visibilities of parts; and the second  is a real valued (( $1$ root $+ 17$ parts) $\times B$ camera views$\times 4$)  dimension vector $\vec{b}$ for the bounding boxes and parts. In both vectors, entries corresponding to invisible parts are set to $0$. 

Denoting $M$ as the dimension of the vector $vec{v}$, and by stacking $vec{v}$ for $N$ occlusion configurations, we can get an $N \times M$ occlusion matrix $\mathcal{D}$, where the first few rows of this matrix for $B=8$ is shown in the right side in Fig.\ref{fig:occlusion}. Note that we have partitioned the view space into $B$ views, so for each row, the visible parts always concentrate in a segment of the vector representing that view. 

In learning an initial And-Or model,  each row in $\mathcal{D}$ corresponds to a small subtree of the root OR node. In particular, each subtree consists of an And-node as the root and a set of
terminal nodes as its children. An example of the data matrix and corresponding initial And-Or model is shown in the middle in Fig.\ref{fig:occlusion}. %Each row of $\mathcal{D}$ represents an occlusion configuration, and each column represents a part. The part is either visible (white), or occluded (gray). 

%\vspace{-1mm}
\subsubsection{Refining the And-Or Structure}\label{sec::compression}
The initial And-Or model is large and redundant, since it has many duplicated occlusion configurations (i.e. duplicated rows in $\mathcal{D}$) and a combinatorial number of part compositions. In the following, we will pursue a compact And-Or structure. The problem can be formulated as:
\begin{equation}\label{eqn::compression}
\min \sum_{i}^{N} \mid v_i - v_i(\mathcal{G})\mid_2^2 + \lambda\mid \mathcal{G} \mid
\end{equation}
where $v_i$ is the $i$-$th$ row of the data matrix $\mathcal{D}$, $v(\mathcal{G})$ returns its most approximate occlusion configuration generated by the And-Or graph (AOG), $|\mathcal{G}|$ is the number of nodes and edges in the structure, and $\lambda$ is the trade-off parameter balancing the model precision and complexity. In each view, we assume the number of occlusion branches is not greater than $K(=4)$.

We solve Eqn.\ref{eqn::compression} using a modified graph compression algorithm similar to \cite{zzAoT}. As illustrated in the right side in Fig.\ref{fig:occlusion}, the algorithm starts from the initial And-Or model, and iteratively combines branches if the introduced loss was smaller than the decrements in complexity term $\lambda |\mathcal{G}|$. This process is equivalent to iteratively finding large blocks of $1$s on the corresponding data matrix through row and column permutations, where an example is shown in the bottom in Fig.\ref{fig:occlusion}. As there are consistently visible parts for each view, the algorithm will quickly converge to the structure shown in  Fig.\ref{fig:demo}.

With the refined And-Or model, we compute occlusion configurations (i.e., the consistently visible parts and optional occluded parts) in each view. In addition, the bounding box size and nominal position of each Terminal-node w.r.t. its parent And-node can also be estimated by geometric means of corresponding values in the vector $\vec{b}$. These information will be used to initialize the latent variables of our model in learning the parameters.

\textbf{Variants of And-Or Models.} 
We will test our model using two types of specifications to be consistent with our two previous conference papers, one is called \textit{And-Or Structure}~\cite{boli_iccv13} for occlusion modeling based on CAD simulation without multi-car context components, and the other called \textit{Hierarchical And-Or Model}~\cite{boli_eccv14} for occlusion and context. We also compare two methods of part selection in hierarchical And-Or model, one is based on the greedy parts as done in the DPM~\cite{DPM}, denoted by \textit{AOG+Greedy},  and the other based on the proposed CAD simulation, denoted by \textit{AOG+CAD}. 

%\vspace{-3mm}
%------------------------------------------------------------------------
\section{Learning Parameters} \label{sec:learning} %\vspace{-2mm}
With the learned And-Or structure, we adopt the WLSSVM method \cite{pffgrammar} in learning the parameters $\Theta=(\Theta^{app}, \Theta^{def}, \Theta^{bias})$ (for appearance, deformation and bias). When the occlusion configurations are mined by CAD simulations (i.e., for the two model specifications, And-Or Structure and AOG+CAD), we will use both the \textit{Step 0} and \textit{Step 1} below in learning parameters, otherwise we use \textit{Step 1} only (i.e., for AOG+Greedy). 

\textbf{Step 0: Initializing Parameters with Synthetic Training Data.}
We learn the initial parameters $\Theta$ with synthetic training data (see Fig.\ref{fig:simulation}). We randomly superimpose the synthetic positive samples on some randomly selected real images without cars appearing (instead of using white background directly, see Fig.\ref{fig:simulation}) to reduce the appearance gap between  the synthetic samples and real car samples.   In the synthetic data, the parse tree $pt$ for each multi-car positive sample is known except that the positions of parts are allowed to deform.  

\textbf{Step 1: Learning Parameters with Real Training Data.}
In the real training data, we only have annotated bounding boxes for single cars.
 The parse tree $pt$ for each multi-car positive sample is hidden except for the multi-car configuration which can be computed based on the annotated bounding boxes of single cars as stated in Sec.\ref{sec:cm}. Then, we initialize the parse tree for each positive sample either based on the initial parameters learned in step 0 (for the And-Or structure and AOG+CAD) or using a similar idea as done in learning the mixture of DPMs [17] to initialize the single-car And-nodes for AOG+Greedy. After the initialization, the parameters $\Theta$ are learned iteratively under the WLSSVM framework. %We initialize the parse tree for each multi-car positive sample based on the initial model learned above. 
 During learning, we run the DP inference to assign the optimal parse trees for multi-car positive samples.  
 
  The objective function to be minimized is defined by,
\begin{equation}
\mathcal{E}(\Theta) = \frac{1}{2} \Arrowvert \Theta \Arrowvert^2 + C \sum_{i=1}^M L'(\Theta, x_i, y_i) \label{eqn:obj}
\end{equation} 
where $x_i\in D_{N-car}^+$ represents a training sample ($N\geq 1$) and $y_i$ is the $N$ bounding box(es). 
$L'(\Theta, x, y)$ is the surrogate loss function,
\begin{align}
\nonumber L'(&\Theta,x,y) =  \max_{pt \in \Omega_{\mathcal{G}}} [score(x, pt; \Theta) + L_{margin}(y, box(pt))]  - \\
&\max_{pt \in \Omega_{\mathcal{G}}} [score(x, pt; \Theta) - L_{output}(y, box(pt))] \label{eqn:loss}
\end{align}
where $\Omega_{\mathcal{G}}$ is the space of all parse trees derived from the And-Or model $\mathcal{G}$, $score(x,pt; \Theta)$ computes the score of a parse tree as stated in Sec.\ref{sec:model}, and $box(pt)$ the predicted bounding box(es) base on the parse tree. As pointed out in \cite{pffgrammar}, the loss $L_{margin}(y, box(pt))$ encourages high-loss outputs to ``pop out" of the first term in the RHS, so that their scores get pushed down.
The loss $L_{output}(y, box(pt))$ suppresses high-loss outputs in the second term in the right hand side, so the score of a low-loss prediction gets pulled up. More details are referred to \cite{pffgrammar,McAllesterLossBound}. In general, since $L'$ in Eqn.(\ref{eqn:loss}) is not convex, the objective function, Eqn.(\ref{eqn:obj}) leads to a nonconvex optimization problem. The WLSSVM adopts the CCCP procedure \cite{CCCP} in optimization, which can find a local optima of the objective. The loss function is defined by,
\begin{equation}
\small 
L_{\ell, \tau}(y, box(pt)) = \left\{ 
  \begin{array}{l l}
    \ell & \quad \text{if $y=\perp$ and pt $\neq \perp$}\\
    0 & \quad \text{if $y=\perp$ and pt $=\perp$}\\
    \ell & \quad \text{if $y\neq \perp$ and $\exists$ $B\in y$} \\ 
    &\qquad \text{with $ov(B, B')< \tau, \forall B'\in box(pt)$}\\
    0 & \quad \text{if $y\neq \perp$ and $ov(B, B')\geq \tau$,} \\ 
    &\qquad \text{$\forall$ $B\in y$ and $\exists B'\in box(pt)$} 
  \end{array} \right.,  
\end{equation}
where $\perp$ represents background output and $ov(\cdot, \cdot)$ is the intersection-union ratio of two bounding boxes. Following the PASCAL VOC protocol we have $L_{margin} = L_{1, 0.5}$ and $L_{output} = L_{\infty, 0.7}$.
 In practice, we modify the implementation in \cite{voc5} for our loss formulation.

%------------------------------------------------------------------------
%\vspace{-2mm}
\section{Experiments}\label{sec:exp} 
In this section, we evaluate our models on four car detection datasets and three car viewpoint estimation dataset and present detail analyses on different aspects of our models. We first introduce two self-collected car datasets of street-parking cars and parking-lot cars respectively (Sec.~\ref{sec:datasets}), and then evaluate the detection performance of our models on four datasets  (Sec.~\ref{sec:detection}): the two self-collected datasets, the KITTI car dataset \cite{Geiger12} and the PASCAL VOC2007 car dataset \cite{pascal}. We further analyze the performance of our model w.r.t. different aspects of our models (Sec.~\ref{sec:analysis}). The performance of car viewpoint estimation is presented in Sec.~\ref{sec:pose}. 

\textit{Training and Testing Time.} In all experiments, we utilize a parallel computing technique to train our model. It takes about 9 hours to train an And-Or Structure model and 16 hours to train a hierarchical And-Or Model due to inferring the assignments of part latent variables on positive training examples and mining hard negatives. For detection, it takes about 2 and 3 seconds to process an image with size of $640 \times 480$ pixels for a And-Or structure and a hierarchical And-Or model, respectively.

%\vspace{-2mm}
\subsection{Datasets} \label{sec:datasets}
To test our model on occlusion and context modeling, we collected two car datasets     
\footnote{http://www.stat.ucla.edu/\textasciitilde boli/publication/street-parking-release.zip and parking\_lot\_release.zip}. 

\begin{figure}
\centering
\includegraphics[width = 0.45\textwidth]{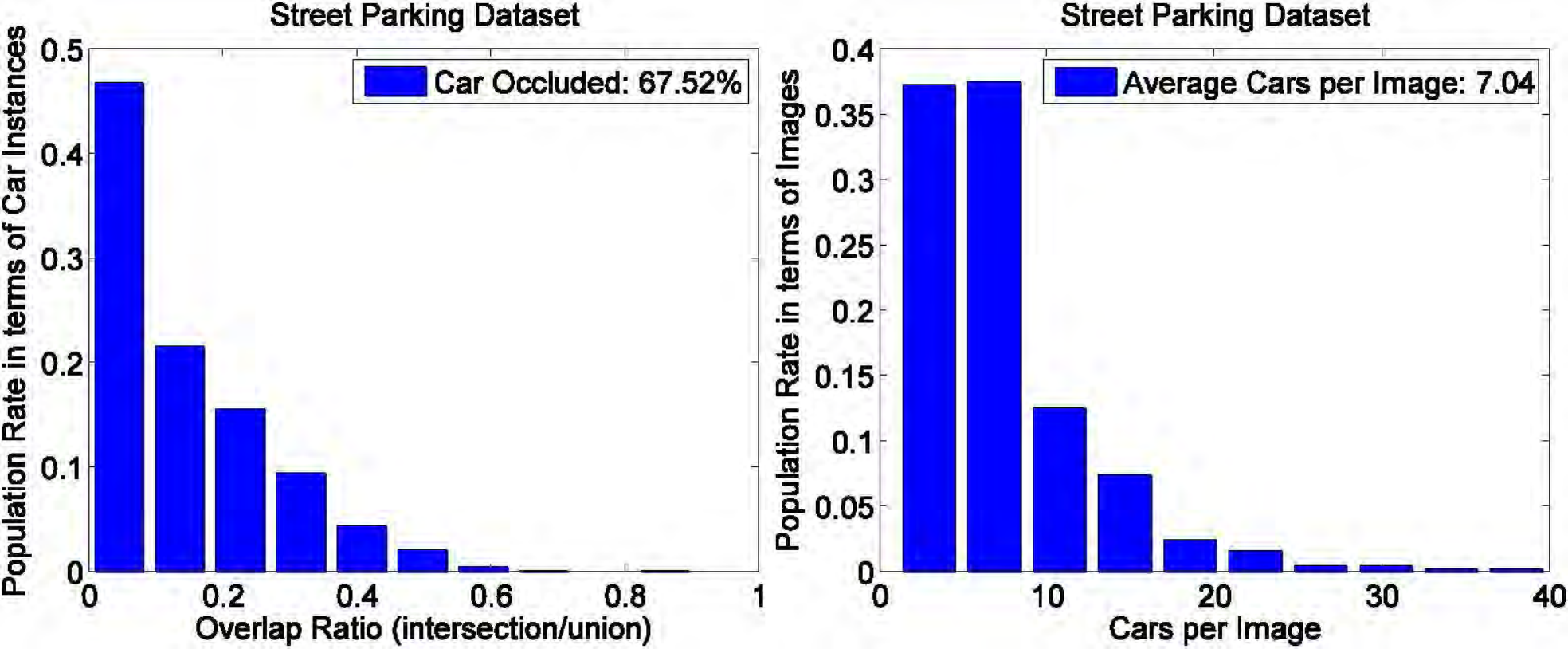}
\resizebox{0.9\hsize}{!}{
\begin{tabular}{|c|c|c|c|}
\hline
{} & Pascal\cite{pascal} & KITTI\cite{Geiger12} & Street Parking \\
\hline
Avg. cars & $1.75$ & $\approx 3$ & $7.04$ \\
\hline
\end{tabular}
}
\caption{Top: The distribution of overlap ratio and cars per image on the Street-Parking dataset. Bottom: Comparison of the average number of cars per image.}
\label{fig:datastats}
\vspace{-4mm}
\end{figure}

\textbf{The Street Parking Car Dataset.} There are several datasets featuring a large amount of car images \cite{eth80,savarese,epfl,pascal}, but they are not suitable to evaluating occlusion handling, as the proportion of (moderately or heavily) occluded cars is marginal. The recently proposed KITTI dataset \cite{Geiger12} contains occluded cars parked along the streets, but it can not fully evaluate the ability of our model since the car views are rather fixed as the video sequences are captured from a car driving on the road (e.g., no birdeye's view). In addition, the average number of cars on each image is still not large enough (mostly $3$ cars, see the statistics in the bottom in Fig.~\ref{fig:datastats}). To provide a more challenging occlusion dataset, we collected one emphasizing street parking cars with heavy occlusions, diverse viewpoint changes and much larger number of cars per image (see the last two rows in Fig.\ref{fig:dets}). The dataset consists of $881$ images. Fig.~\ref{fig:datastats} shows the bounding box overlapping distribution and average number of cars per image. For the simplicity of annotation, we only label the bounding boxes of single cars in each image. We split the dataset into training and testing sets containing $440$ and $441$ images, respectively.

\textbf{The Parking Lot Dataset.}  Our Street Parking Car Dataset provides more viewpoints, however, the context and occlusion configurations are relatively restricted (most cars just compose the head-to-head occlusions). To thoroughly evaluate our models in terms of both context and occlusions, we collected the parking lot car dataset, which has larger occlusion variations and larger number of cars in each image (see the $4$-th and $5$-th rows in Fig. \ref{fig:dets}). It contains $65$ training images and $63$ testing images. Although the number of images is small, the number of cars is noticeably large, with $3,346$ cars (including left-right mirrored ones) for training and $2,015$ cars for testing. 
\subsection{Detection} \label{sec:detection}
We test our hierarchical And-Or Model on four challenging datasets. %For simplicity, we just use greedy part selection for occlusion modelling.

%\vspace{-2mm}
\subsubsection{Results on the KITTI Dataset}
\begin{figure}[!t]
\begin{center}
\includegraphics[width = 0.42\textwidth]{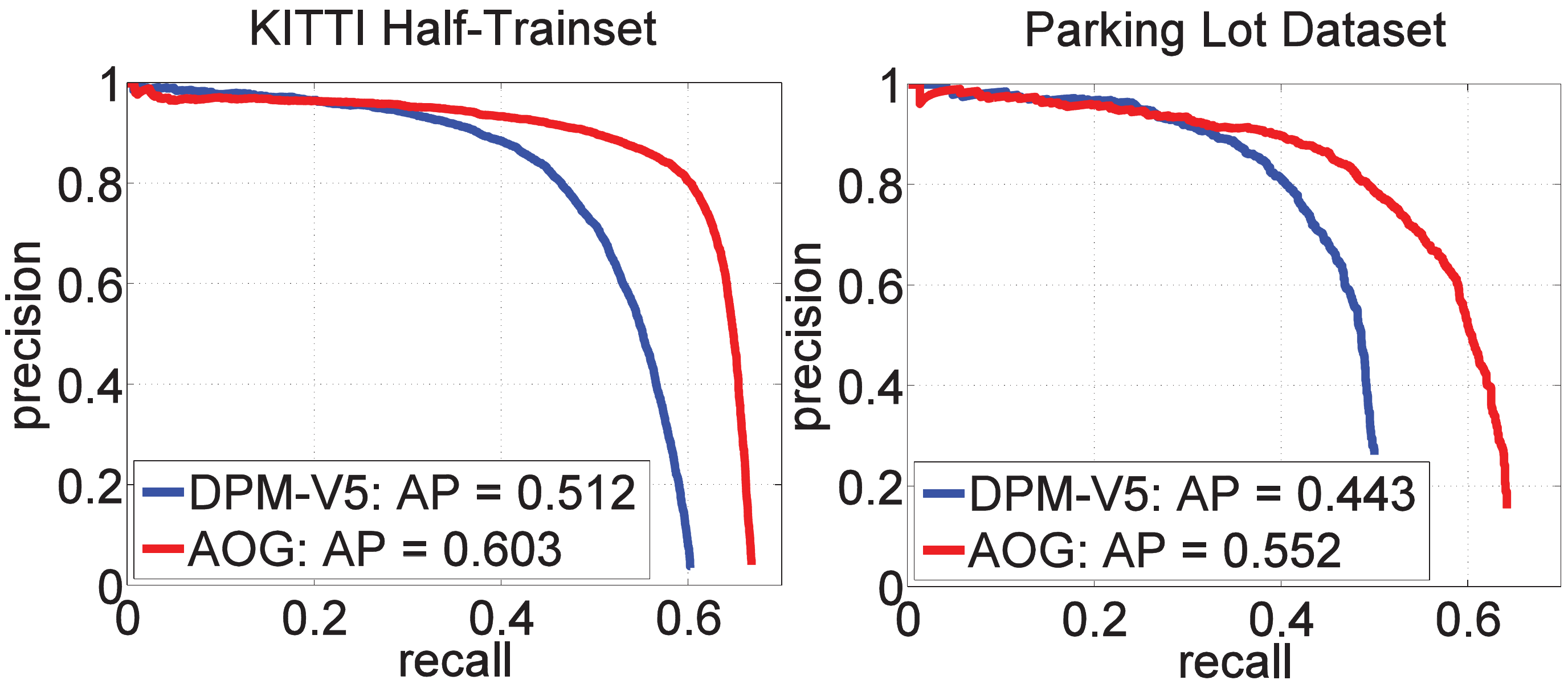} 
\caption{Precision-recall curves on the test subset splitted from the  KITTI  trainset (Left) and the Parking Lot dataset (Right).}
\label{fig:pr_results}
\end{center}
\vspace{-4mm}
\end{figure}
 
The KITTI dataset \cite{Geiger12} contains $7,481$ training images and $7,518$ testing images, which are captured from an autonomous driving platform. We follow the provided benchmark protocol for evaluation.
Since the authors of \cite{Geiger12} have not released the test annotations, we test our model in the following two settings.

\textbf{Training and Testing by Splitting the Trainset.}
We randomly split the KITTI trainset into the training and testing subsets equally. 

\textit{Baseline Methods.} Since DPM \cite{DPM} is a very competitive model with source code publicly available, we compare our model with the latest version of DPM
(i.e., voc-release5 \cite{voc5}). The number of components are set to $16$ as the baseline methods trained in \cite{Geiger12}, other parameters are set as default.

\textit{Parameter Settings.} We consider multi-car contextual patterns with the number of cars $N=1, 2$. We set the number of context patterns and occlusion configurations to be $10$ and $16$, respectively. As a result, the learned hierarchical And-Or model has $10$ $2$-car configurations in layer $1$, and $16$ single car branches in layer $3$ (see Fig. \ref{fig:demo}). 

\begin{table}
\begin{center}
\resizebox{1.0\hsize}{!}{
\begin{tabular}{|l|l|l|l|}
%\hline
%\multicolumn{4}{|c|}{Test On KITTI Benchmark \cite{Geiger12}} \\
\hline
Methods & Easy & Moderate & Hard \\
\hline
mBow \cite{behley2013iros} & $36.02\%$ & $23.76\%$ & $18.44\%$ \\
\hline
LSVM-MDPM-us \cite{DPM} & $66.53\%$ & $55.42\%$ & $41.04\%$ \\
\hline
LSVM-MDPM-sv \cite{DPM,Geiger11} & $68.02\%$ & $56.48\%$ & $44.18\%$ \\
\hline
MDPM-un-BB \cite{DPM} & $71.19\%$ & $62.16\%$ & $48.43\%$ \\
\hline
OC-DPM \cite{bojan_cvpr13} & $74.94\%$ & $65.95\%$ & $53.86\%$ \\
\hline
DPM \cite{voc5} (trained by us) & $77.24\%$ & $56.02\%$ & $43.14\%$ \\
\hline
MV-RGBD-RF \cite{mv_rgbd} & $76.40\%$ & $69.92\%$ & $57.47\%$ \\
\hline
SubCat \cite{OhnBar} & $84.14\%$ & $75.46\%$ & $59.71\%$ \\
\hline
3DVP \cite{xiang_cvpr15} & \underline{$\mathbf{87.46\%}$} & $75.77\%$ & \underline{$\mathbf{65.38\%}$} \\
\hline
Regionlets \cite{regionlets} & $84.75\%$ & \underline{$\mathbf{76.45\%}$} & $59.70\%$ \\
\hline
AOG+Greedy-Half & $84.36\%$ & $71.88\%$ & $59.27\%$ \\
\hline
AOG+Greedy-Full & ${84.80\%}$ & ${75.94\%}$ & ${60.70\%}$ \\
\hline
\end{tabular}
}
\end{center}
\caption{Performance comparison (in AP) on the KITTI benchmark \cite{Geiger12}.}\label{tab:bench}
\vspace{-3mm}
\end{table}
\begin{table}
\begin{center}
\resizebox{1.0\hsize}{!}{
\begin{tabular}{|c|c|c|c|c|}
%\hline
%\multicolumn{5}{|c|}{Test On the Street Parking Dataset \cite{boli_iccv13}} \\
\hline
{} & DPM \cite{voc5} & And-Or Structure \cite{boli_iccv13} & AOG+Greedy & AOG+CAD\\
\hline
AP & $52.0\%$ & $57.8\%$ & $62.1\%$ & $\mathbf{65.3\%}$ \\
\hline
\end{tabular}
}
\end{center}
\caption{Performance comparison (in AP) on the Street Parking dataset \cite{boli_iccv13}.}\label{tab:street}
\vspace{-4mm}
\end{table}

\textit{Detection Results.}
The left figure in Fig. \ref{fig:pr_results} shows the precision-recall curves of DPM and our model. Our model outperforms DPM by $9.1\%$ in terms of average precision (AP). The performance gain comes from both precision and recall, which shows the importance of context and occlusion modeling.

\textbf{Testing on the KITTI Benchmark.}
We evaluate our model with two different training data settings: one trained using half training set on the KITTI testset, denoted by AOG+Greedy-Half, and the other trained with full training set, denoted by AOG+Greedy-Full (which has $16$ context patterns and $32$ occlusion configurations).

The benchmark has three subsets (\emph{Easy, Moderate, Hard}) w.r.t the difficulty of object size, occlusion and truncation. All methods are ranked based on performance in the moderately difficult subset. Our entry in the benchmark is ``AOG". 
Table \ref{tab:bench} shows the detection results of our model and other state-of-the-art models. Here, we omit the CNN-based method, as they are all anonymous submissions. Details of the benchmark results are available at \textit{http://www.cvlibs.net/datasets/kitti/eval\_object.php}.

 Our AOG+Greedy-Full outperforms all the DPM-based models. Compared with their best model, OC-DPM \cite{bojan_cvpr13}, our model improved performance on the three subsets by $9.86\%$, $9.99\%$, and $6.84\%$ respectively. We also compare with the baseline DPM trained by ourselves using the voc-release5 code \cite{voc5}, and obtain $7.56$, $19.92\%$ and $17.56\%$ performance gains on the three stubsets. For other DPM based methods trained by the benchmark authors, our model outperforms the best one - MDPM-un-BB by $13.61\%$, $13.78\%$ and $12.27\%$ respectively.

Our model is comparable with SubCat \cite{OhnBar}, 3DVP \cite{xiang_cvpr15} and Regionlets \cite{regionlets}. We achieve slightly better performance than Regionlets \cite{regionlets} on the \emph{Easy} and \emph{Hard} sets, but lose a bit AP on the \emph{Moderate} set. Though our method obtains better rank than 3DVP \cite{xiang_cvpr15} on the moderately difficult set, it performs slightly worse on the easy and hard subsets, which shows the promise of 3D occlusion modeling and subcategory clustering \cite{OhnBar,xiang_cvpr15}. %In this paper, we focus on object modelling and representation, and our model achieves the best HOG feature only results on this benchmark, while \cite{regionlets,xiang_cvpr15,OhnBar} using different features. So our work is complementary with them.

Comparing AOG+Greedy-Half and AOG+Greedy-Full, we can observe that the major  improvement ($4.06\%$) of AOG+Greedy-Full comes from the \emph{Moderate} set, while on the \emph{Easy} and \emph{Hard} sets, we obtain small improvement ($0.44\%$ and $1.43\%$, respectively). These results meet some analyses in \cite{ramanan_modedata}, which indicate there are still large potential improvement on object representation, and much effort should be devoted to improving our current hierarchical And-Or model.

The first $3$ rows in Fig. \ref{fig:dets} show the qualitative results of our model. The red bounding boxes show  successful detection, the blue ones  missing detection, and the green ones false alarms. In experiments, our model is robust to detect cars with heavy car-to-car occlusions and background clutters. The failure cases are mainly due to extreme occlusions, extremly low resolution, large car deformation and/or inaccurate (or multiple) bounding box localization.

% \begin{figure*}[!t]
% \centering
% \includegraphics[width = 0.9\textwidth]{./dets22.pdf}
% \caption{Examples of successful and failure cases by our model on the KITTI dataset (first 3 rows), the Parking Lot dataset (the $4$-th and $5$-th rows) and the Street Parking dataset (the last two rows).
% Best viewed in color and magnification.}
% \label{fig:dets} 
% \vspace{-4mm}
% \end{figure*}

\begin{SCfigure*}
  %\centering
  \caption{Examples of successful and failure cases by our model on the KITTI dataset (first 3 rows), the Parking Lot dataset (the $4$-th and $5$-th rows) and the Street Parking dataset (the last two rows).
  Best viewed in color and magnification.}\label{fig:dets} \vspace{-3mm}
  \includegraphics[width=0.8\textwidth]%
    {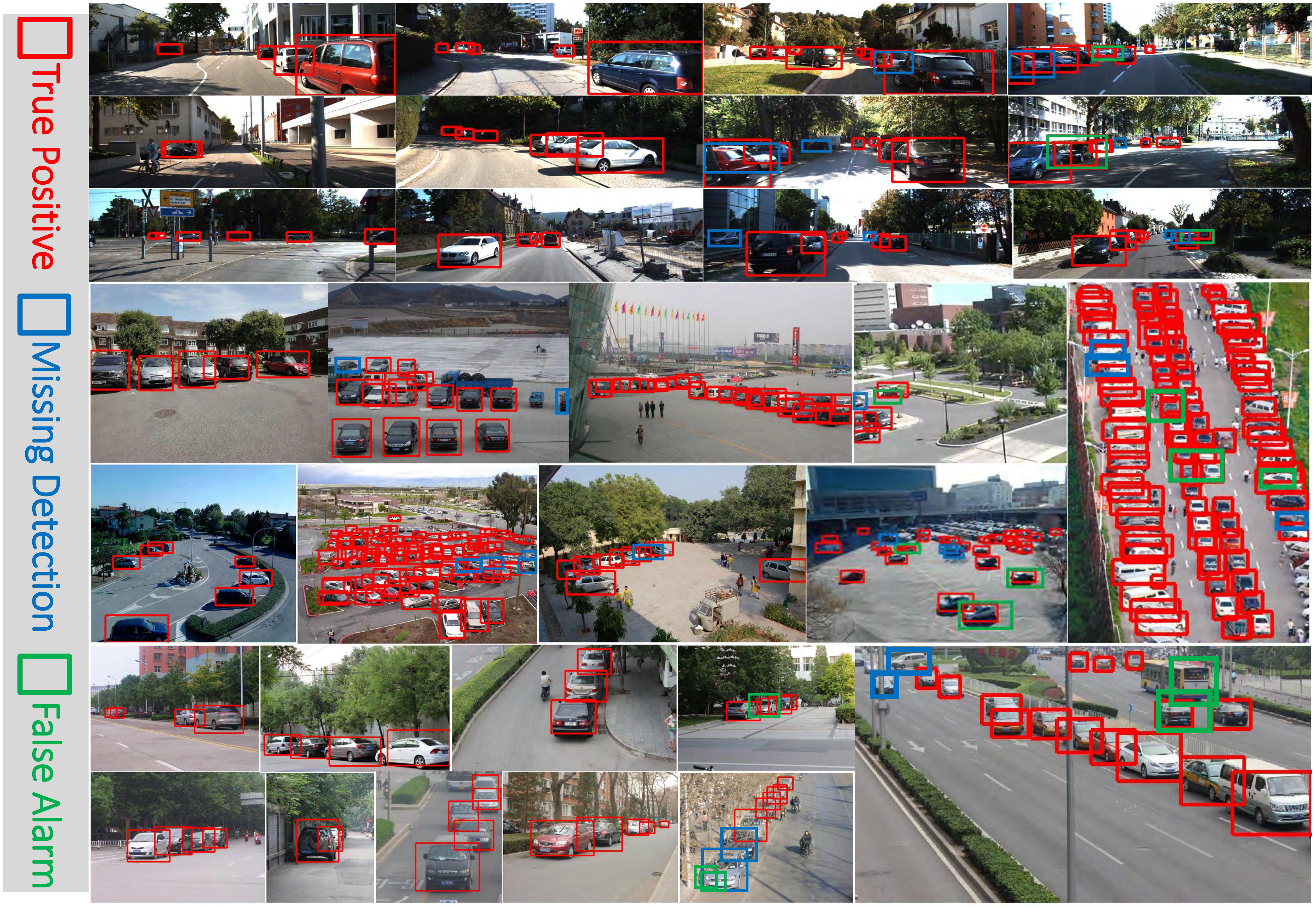}% picture filename
\end{SCfigure*}

%\vspace{-2mm}
\subsubsection{Results on the Parking Lot Dataset}  
\textit{Evaluation Protocol.}
We follow the PASCAL VOC evaluation protocol \cite{pascal} with the overlap of intersection over union being greater than or equal to $60\%$ (instead of original $50\%$). In practice, we set this threshold to make a compromise between localization accuracy and detection difficulty. 
The detected cars with bounding box height smaller than $25$ pixels do not count as false positives as done in \cite{Geiger12}. We compare with the latest version of DPM implementation \cite{voc5} and set the number of contextual patterns and occlusion configurations to be $10$ and $18$ respectively.

\textit{Detection Results.}
 The right side in Fig. \ref{fig:pr_results} shows the performance comparisons between our model and DPM. Our model obtains $55.2\%$ in AP, which outperforms the latest version of DPM by $10.9\%$.
The fourth and fifth rows in Fig. \ref{fig:dets} show the qualitative results. Our model is capable of detecting cars with different occlusions and viewpoints. 

%\vspace{-2mm}
\subsubsection{Results on the Street Parking Dataset} 
To compare with the benchmark methods, we follow the evaluation protocol provided in  \cite{boli_iccv13}.

Results of our model and other benchmark methods are shown in Table \ref{tab:street}, our hierarchical And-Or model outperforms DPM \cite{voc5} and our previous And-Or Structure \cite{boli_iccv13} by $10.1\%$ and $4.3\%$ respectively. We think the performance is improved due to the joint representation of context patterns and occlusion configurations.
The last two rows in Fig. \ref{fig:dets}  show some qualitative examples. Our model is capable of detecting occluded street-parking cars, meanwhile it also has a few inaccurate detection results and misses some cars (mainly due to low resolution). 

%\vspace{-2mm}
\subsection{Diagnosing the Performance of our Model} \label{sec:analysis}
In this section, we evaluate various aspects to diagnose the effects of each individual component in our model.

%\vspace{-2mm}
\subsubsection{The Effect of Occlusion Modeling}
Our And-Or Structure model is based on CAD simulation. Thus in the first analysis, we test the effectiveness of the learned And-Or structure in representing different occlusion configurations. To this purpose, we generate a synthetic dataset using 5,040 $3$-car synthetic images as our training data, and a mixture of 3,000 $3$-car and $7$-car (placed in a $1 \times 7$ grid) synthetic images as our testing data. For each generated image, we add the background from the category \emph{None} of the TU Graz-02 dataset \cite{tu} and apply Gaussian blur to reduce the boundary effects. Samples of the training and testing data are shown on the left and middle in Fig.\ref{fig:simulation}. In experimental comparisons, the best DPM has $16$ components and the best And-Or structure has $8$ views with $19$ occlusion configurations, $5$ layers and $111$ nodes in total. As shown in the right side in Fig.\ref{fig:simulation}, our model outperforms the DPM by 7.2\% in AP. 

\begin{figure}
\centering
\includegraphics[width = 0.5\textwidth]{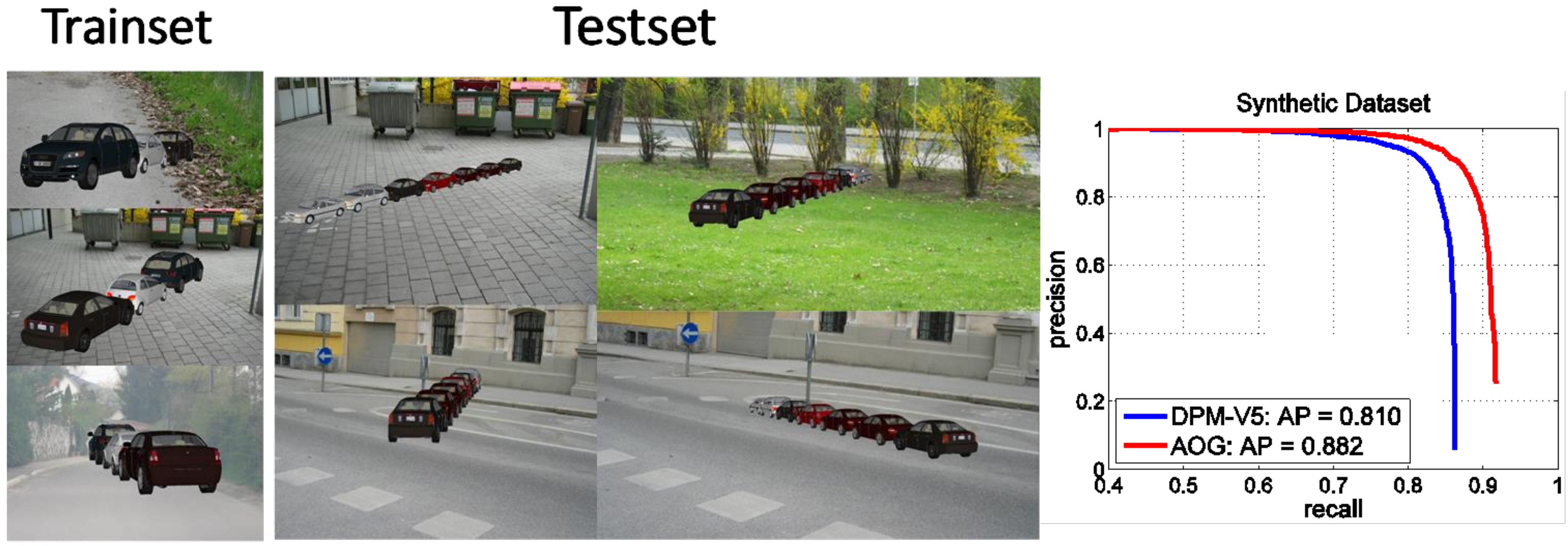}
\caption{Left and Middle: Training and testing samples from the synthetic dataset. Right: detection results of DPM and And-Or Structure.}
\label{fig:simulation} 
\vspace{-3mm}
\end{figure}

\begin{figure*}%[!t]
\centering
\includegraphics[width = 1.0\textwidth]{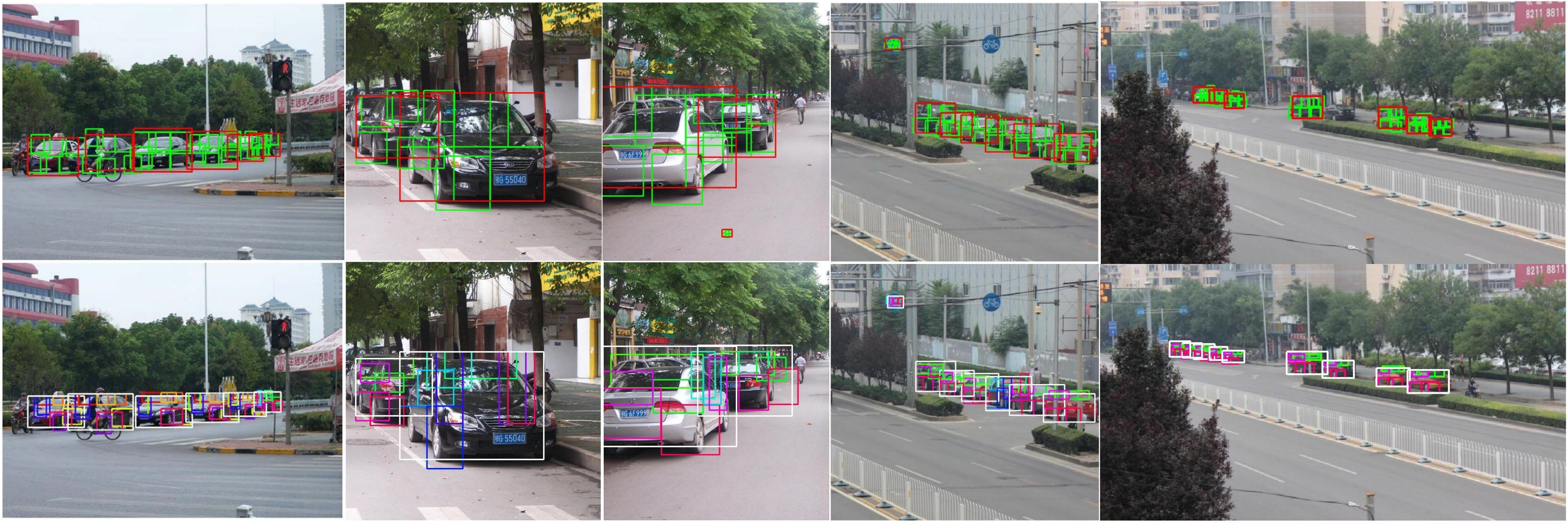} 
\caption{Visualization of part layouts output by our AOG+Greedy (Top) and AOG+CAD (Bottom). Best viewed in color and magnification.}
\label{fig:parts} 
%\vspace{-3mm}
\end{figure*}

%\vspace{-2mm}
\subsubsection{The Effect of CAD Simulation in Real Situations}
To verify the effectiveness of our And-Or Structure model in terms of occlusion modeling, we compare it with state-of-the-art DPM \cite{DPM}. Both of these two models are based on part-level occlusion modeling. The And-Or Structure learns  semantic visible parts based on CAD simulations. The DPM handles occlusion implicitly by introducing a trunction feature at each HOG cell. 
The second and third column in Table \ref{tab:street} show their performance on  Street Parking dataset. We can see the semantic visible parts learned from CAD simulations can generalize to real datasets.
By adding context, we are interested in whether it affects the effectiveness of occlusion modeling. To compare AOG+Greedy and AOG+CAD fairly, 
they have the same number of context patterns and occlusion configurations,  $8$ and $16$ respectively. As shown in the fourth and fifth column in Table \ref{tab:street}, AOG+CAD performs better than AOG+Greedy, which shows the advantage of modeling occlusion using semantic visible parts. 

Fig. \ref{fig:parts} shows the inferred part bounding boxes by AOG+Greedy and AOG+CAD. We can observe that the semantic parts in AOG+CAD are meaningful, although they may be not accurate enough in some examples.

\begin{table}
\begin{center}
\resizebox{1\hsize}{!}{
\begin{tabular}{|c|c|c|c|}
%\hline
%\multicolumn{5}{|c|}{Test On the Street Parking Dataset \cite{boli_iccv13}} \\
\hline
{car} & DPM \cite{voc5} & And-Or Structure \cite{boli_iccv13} & AOG+Greedy\\
\hline
AP & $58.2\%$ & $58.7\%$ & $\mathbf{60.6\%}$ \\
\hline
\end{tabular}
}
\end{center}
\caption{Performance comparison (in AP) on the PASCAL VOC 2007\cite{pascal}.}\label{tab:pascal}
\vspace{-5mm}
\end{table}

%\vspace{-2mm}
\subsubsection{The Effect of Multi-car Context Modeling}
The state-of-the-art models are mainly based on single car modeling. To evaluate the effectiveness of context, we compare our hierarchical And-Or model with other non-context models in Table \ref{tab:bench}. We can see that our model outperforms all other models in different occlusion settings. Specifically, our model outperforms DPM by a large margin (above 10\% in AP) on the ``Moderate" and ``Hard" KITTI test data, which shows context is very important to object detection especially in heavily occluded car-to-car situations.

On the Street Parking dataset, we observe the same results. In Table \ref{tab:street}, both AOG+Greedy and AOG+CAD outperform DPM and And-Or Structure by a large margin. Here, AOG+Greedy and AOG+CAD jointly model context and occlusions, while DPM and And-Or Structure model occlusions only.

%\vspace{-2mm}
\subsubsection{Performance on General Occlusion Settings}
Our model is generalizable in terms of context and occlusion modeling, it can cope with both occlusion and non-occlusion situations. To verify our model on less occluded settings, we use the PASCAL VOC 2007 Car dataset as a testbed. As analyzed by Hoiem, et. al. in \cite{HoiemDiagnoseError}, cars in the PASCAL VOC dataset do not have much occlusions and car-to-car context.

We first show that our And-Or Structure is capable to detect cars on the PASCAL VOC 2007 as well as the DPM method \cite{voc5}.
To approximate the occlusion configurations observed on this dataset, we generate synthetic images with car-to-car occlusions and car self-occlusions. For the car-to-car occlusions, we use the full $3 \times 3$ grid instead of the special case in the street parking dataset. Correspondingly, the learned And-Or structure contains branches for self-occlusions as well as those for car-to-car occlusions. On this dataset, the DPM has $6$ components and the And-Or structure has $6$ views with $10$ occlusion configurations, $5$ layers and $109$ nodes.

The third column  in Table \ref{tab:pascal} shows the performance of our And-Or structure model and the DPM. Our model achieves slightly better recall than DPM, which meets the analysis in \cite{HoiemDiagnoseError}. This experiment shows that our And-Or structure method does not lose performance in general datasets.

Then, we verify our hierarchical And-Or model is capable to detect cars on the PASCAL VOC 2007 as well as other single object models. We compare with the latest version of DPM \cite{voc5}. The APs are 60.6\% (our model) and 58.2\% (DPM) respectively (Table \ref{tab:pascal}).

\begin{table}
\begin{center}
\resizebox{0.7\hsize}{!}{
\begin{tabular}{|l|l|l|l|l|l|}
\hline
\multicolumn{6}{|c|}{Pascal VOC 2006 Car Dataset\cite{pascal}} \\
\hline
{} & DPM & \cite{Lopez2011} & \cite{gu10} & \cite{sun09} & ours \\
\hline
MPPE & $0.69$ & $0.73$ & $\mathbf{0.86}$ & $0.57$ & $0.73$ \\
\hline
\end{tabular}
}
\vspace{2mm}

\resizebox{1\hsize}{!}{
\begin{tabular}[width = 1in]{|l|l|l|l|l|l|l|l|}
\hline
\multicolumn{8}{|c|}{3D Car Dataset \cite{savarese}} \\
\hline
{} & DPM & \cite{Lopez2011} & \cite{Liebelt10} & \cite{glasner} & \cite{teach3D}$^1$ & \cite{teach3D}$^2$ & ours \\
\hline
AP & $99.6$ & $96$ & $76.7$ & $99.2$ & $\mathbf{99.9}$ & $99.7$ &$\mathbf{99.9}$ \\
\hline
MPPE & $86.3$ & $89$ & $70$ & $85.3$ & $\mathbf{97.9}$ & $96.3$ & $94$ \\
\hline
\end{tabular}
}
\end{center}
\caption{View Estimation on Pascal VOC 2006 Car Dataset \cite{pascal} and 3D Car Dataset \cite{savarese}. \cite{teach3D}$^1$ and \cite{teach3D}$^2$ refer to DPM-VOC+VP and DPM-3D-Constraints, respectively.}\label{view_estimation}
\vspace{-4mm}
\end{table}

\begin{table*}
\begin{center}
\resizebox{0.9\hsize}{!}{
\begin{tabular}{|c|c|c|c|c|c|}
%\hline
%\multicolumn{5}{|c|}{Test On the Street Parking Dataset \cite{boli_iccv13}} \\
\hline
{} & VDPM \cite{xiang_wacv14} & DPM-VOC+VP \cite{teach3D} & (fisher+spm) \cite{pedersoli} & (decaf) \cite{pedersoli} & our And-Or Structure\\
\hline
4 views & $37.2\% / 20.2\%$ & $45.6\% / 36.9\%$ & $36.1\% / 28.9\%$ & $36.1\% / 24.1\%$ & $43.0\% / 34.3\%$ \\
\hline
8 views & $37.3\% / 23.5\%$ & $47.6\% / 36.6\%$ & $36.1\% / 26.6\%$ & $36.1\% / 23.3\%$ & $44.9\% / 33.2\%$ \\
\hline
16 views & $36.6\% / 18.1\%$ & $46.0\% / 29.6\%$ & $36.1\% / 19.6\%$ & $36.1\% / 19.4\%$ & $43.2\% / 27.6\%$ \\
\hline
24 views & $36.3\% / 13.7\%$ & $42.1\% / 24.6\%$ & $36.1\% / 15.9\%$ & $36.1\% / 16.7\%$ & $41.1\% / 22.9\%$ \\
\hline
\end{tabular}
}
\end{center}
\caption{The results of VDPM, DPM-VOC+VP and And-Or Structure on the PASCAL3D+ Car Dataset \cite{xiang_wacv14}. The first number indicates the average precision (AP) for detection and the second number shows the average viewpoint precision (AVP) for joint object detection and view estimation.}\label{tab:pascal3d}
\vspace{-5mm}
\end{table*}

%\vspace{-3mm}
\subsection{View Estimation} \label{sec:pose}
With the help of CAD simulations, our And-Or Structure model can compute the viewpoints of detected cars. To verify the capability of view estimation,  we perform $2$ experiments.

Firstly, we report the mean precision in pose estimation (MPPE), equivalent to the means of confusion matrix diagonals, on both the Pascal VOC 2006 car dataset \cite{pascal-voc-2006} and the 3D Object dataset \cite{savarese}.
The 3D Object Classes dataset \cite{savarese} is introduced in 2007. For each class, it has images of 10 different object instances with 8 different poses. We follow the evaluation protocol described in \cite{savarese}: 7 randomly selected car instances are used for training, and 3 instances for testing. The 2D car bounding boxes are computed from the annotated segmentation masks. The negative examples are collected from the PASCAL VOC 2007 car dataset.
For the VOC 2006 car database \cite{pascal-voc-2006}, there are 469 cars with viewpoint labels (frontal, rear, left and right). We only use these labeled images with the standard training/test split. The detection performance is evaluated through precision-recall (PR) curve.
For view estimation, the two datasets emphasize visible cars. Our And-Or structure has $8$ views with $8$ (self-occlusion) branches, $5$ layers and $90$ nodes.
Table \ref{view_estimation} shows the comparison of our model with the state-of-the-art methods on these two datasets. Our model is comparable to or better than some recently proposed models \cite{Lopez2011,gu10,teach3D}.

Secondly, we compare our model with the state-of-the-art models on the recently proposed PASCAL3D+ Dataset \cite{xiang_wacv14}. This dataset augments $12$ rigid categories in the PASCAL VOC 2012 \cite{pascal} with 3D annotations by fitting CAD models with 2D images semi-manually. It is a challenging dataset for 3D object detection and pose estimation.  We test on the car category. We use the metric - Average Viewpoint Precision (AVP) \cite{xiang_wacv14} to simultaneously evaluate 2D bounding box localization and viewpoint estimation. In computing the AVP, a candidate detection is considered to be a true positive if and only if the bounding box overlap is larger than $50\%$ and the viewpoint is correct.

Table~\ref{tab:pascal3d} shows the results of our model and the state-of-the-art methods. Our method is better than VDPM \cite{xiang_wacv14} and a deep-cnn-feature-based model (decaf) \cite{pedersoli}. Our And-Or Structure is comparable with \cite{teach3D}, which also used CAD models to learn viewpoints and part-level car geometry.

%\vspace{-3mm}
%------------------------------------------------------------------------
\section{Conclusion} \label{sec:conclusion}
In this paper, we present an And-Or model to represent context and occlusion for car detection and viewpoint estimation. 
The model structure is learned by mining multi-car contextual patterns and occlusion configurations at three levels: a) multi-car layouts, b) single car and c) parts.
Our model is organized in a directed and acyclic graph structure so the efficient DP algorithm can be used in inference. The model parameters are learned by WLSSVM\cite{pffgrammar}. Experimental results show that our model is effective in modeling context and occlusion information in complex situations, and achieves better performance over state-of-the-art car detection methods and comparable performance on viewpoint estimation. 

There are two main limitations in our current implementation. The first one is that we exploited the multi-car contextual patterns using $2$-car composite only. In the scenarios similar to street parking cars and parking lot cars, we could explore multi-car context with more than 2 spatially-aligned cars, as well as 3D scene parsing context~\cite{xiaobai_3DScene}. The second one is that we utilized only the HOG features for appearance. Based on the recent progress on feature learning by convolutional neural network (CNN) \cite{CNN_Hinton, RCNN}, we can also substitute the HOG by the CNN features. Both aspects are addressed in our on-going work and may potentially improve the performance. 

Meanwhile, we are applying the proposed method to other object categories and studying  different ways of mining contextual patterns and occlusion configurations (e.g., integrating with the And-Or quantization methods for 2D object modeling \cite{xisong_cvpr} and 3D car modeling  \cite{Wenze3DCar}).

\ifCLASSOPTIONcompsoc
  \section*{Acknowledgments}
\else
  \section*{Acknowledgment}
\fi

B. Li is supported by China 973 Program under Grant no. 2012CB316300. T.F. Wu and S.C. Zhu are supported by DARPA MSEE project FA 8650-11-1-7149, MURI grant ONR N00014-10-1-0933, and NSF IIS1018751. We thank Dr. Wenze Hu for helpful discussions.

\ifCLASSOPTIONcaptionsoff
  \newpage
\fi



\bibliographystyle{IEEEtran}
\bibliography{AOGcontextocc}

%\begin{thebibliography}{1}


%\end{thebibliography}


\begin{IEEEbiography}[ {
		\includegraphics*[width=.9in,clip]{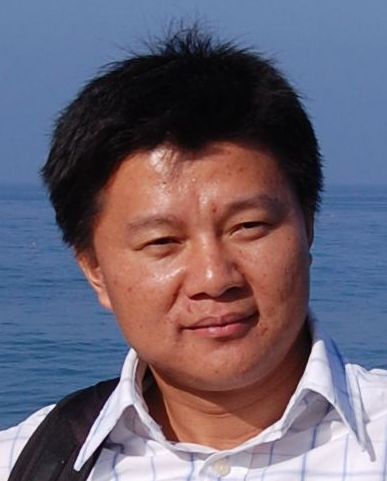} } ]{Tianfu Wu}
	received a Ph.D. degree in Statistics from University of California, Los Angeles (UCLA) in 2011.
	He is currently a research assistant professor in the center for vision, cognition, learning and autonomy (VCLA)  at UCLA. His research interests include: (i) Statistical learning of large scale hierarchical and compositional models (e.g., And-Or graphs) from images and videos. (ii) Statistical inference by learning near-optimal cost-sensitive decision policies. (iii) Statistical theory of performance guaranteed learning algorithm and inference procedure.
\end{IEEEbiography} 

\begin{IEEEbiography}[ {
\includegraphics*[width=.9in,clip]{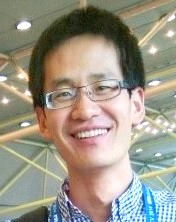} } ]{Bo Li}
received a B.S. degree from Beijing Institute of Technology in 2010. He is currently a Ph.D. student in School of Computer Science and Technology, Beijing Institute of Technology, and a visiting student at Center for Vision, Cognition, Learning and Autonomy (VCLA) at the University of California, Los Angeles (UCLA). His research interests are in pattern recognition, machine learning and computer vision, with a focus on car detection in terms of both 2D and 3D models.
\end{IEEEbiography}


%\newpage


\begin{IEEEbiography}[ {
\includegraphics*[width=.9in,clip]{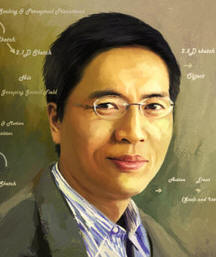} } ]{Song-Chun Zhu}
received a Ph.D. degree from Harvard University in 1996.
He is currently a professor of Statistics and Computer Science at
UCLA, and the director of the Center for Vision, Cognition, Learning and Autonomy.
 He has published over 160 papers in computer vision,  statistical
modeling and learning, cognition, and visual arts.
He received a number of honors, including the J.K. Aggarwal prize
from the Int'l Association of Pattern Recognition in 2008 for
"contributions to a unified foundation for visual pattern
conceptualization,  modeling,  learning, and inference",
 the David Marr Prize in 2003 with Z. Tu et al. for image parsing,
twice Marr Prize honorary nominations in 1999 for texture modeling and
in 2007 for object modeling with Z. Si and Y.N. Wu. He received the
Sloan Fellowship in 2001, a US NSF Career Award in 2001, and an US ONR
Young Investigator Award in 2001.  He received the Helmholtz Test-of-time award in ICCV 2013, and he is a Fellow of IEEE since 2011. 
\end{IEEEbiography}


\vfill


\end{document}